\documentclass{article}

\PassOptionsToPackage{numbers, compress}{natbib}

 \usepackage[dandb, final]{neurips_2025}

\usepackage[utf8]{inputenc} 
\usepackage[T1]{fontenc}    
\usepackage{hyperref}       
\usepackage{url}            
\usepackage{booktabs}       
\usepackage{amsfonts}       
\usepackage{nicefrac}       
\usepackage{microtype}      
\usepackage{xcolor}         

\usepackage{graphicx}
\usepackage{amsmath}
\usepackage{amssymb}
\usepackage{pifont}
\usepackage{makecell}
\usepackage{algorithm}
\usepackage{algpseudocode}
\usepackage{tikz}
\usepackage[accsupp]{axessibility} 
\usepackage{multirow, array}
\usepackage{tabularx, tabulary, booktabs, colortbl}
\usepackage[overload]{ragged2e}
\usepackage{makecell}
\usepackage[most]{tcolorbox}  
\usepackage{subcaption}
\usepackage{xspace}
\usepackage{float}
\usepackage{enumitem}
\usepackage{wrapfig}

\newcommand{\cmark}{\ding{51}}%
\newcommand{\xmark}{\ding{55}}%

\definecolor{lightblue}{rgb}{0.1,0.3,1}
\definecolor{lightgreen}{rgb}{0.1,0.8,0.1}
\definecolor{defaultcolor}{gray}{0.9}
\newlength\savewidth\newcommand\shline{\noalign{\global\savewidth\arrayrulewidth
  \global\arrayrulewidth 1pt}\hline\noalign{\global\arrayrulewidth\savewidth}}
\newcommand{\tablestyle}[2]{\setlength{\tabcolsep}{#1}\renewcommand{\arraystretch}{#2}\centering\footnotesize}

\newcommand{\eat}[1]{\ignorespaces}

\newcommand{\dataset}{\textsc{MedicalNarratives}\xspace}
\newcommand{\model}{\textsc{GenMedCLIP}\xspace}
\newcommand{\modelthreetwo}{\textsc{GenMedCLIP-32}\xspace}
\newcommand{\modelPMB}{\textsc{GenMedCLIP-PMB}\xspace}

\newcommand{\pubmedclip}{\textsc{PubMedCLIP}\xspace}

\newcommand{\countlen}{\textsc{4.7}\xspace}
\newcommand{\pmcclip}{\textsc{PMC-CLIP}\xspace}
\newcommand{\biomed}[0]{\textsc{BiomedCLIP}\xspace}

\title{\dataset: Connecting Medical Vision and Language with Localized Narratives}

\author{ \bf Wisdom O. Ikezogwo $^{1*}$ \: \:
\bf Kevin Zhang  $^{1*}$ \: \: \\
\bf Mehmet Saygin Seyfioglu $^{1,3}$ \: \:
\bf Fatemeh Ghezloo $^{1}$ \: \:
\bf Linda Shapiro $^{1}$  \: \:
\bf Ranjay Krishna $^{1,2}$  \: \: \\
$^{1}$ University of Washington \\
$^{2}$ Allen Institute for Artificial Intelligence \\
$^{3}$ Amazon
}

\begin{document}
\maketitle

\begin{abstract}

Multi-modal models are data hungry. While datasets with natural images are abundant, medical image datasets can not afford the same luxury. 
To enable representation learning for medical images at scale, we turn to YouTube, a platform with a large reservoir of open-source medical pedagogical videos.
We curate MedicalNarratives, a dataset 4.7M medical image-text pairs, with 1M samples containing dense annotations in the form of spatial traces (and bounding boxes), and 118K videos centered on the trace event (with aligned text), enabling spatiotemporal grounding beyond single frames.
Similar to \textit{think-aloud} studies where instructors speak while hovering their mouse cursor movements over relevant image regions, 1M images in MedicalNarratives contains localized mouse traces in image pixels, creating a spatial and temporal association between the text and pixels.
To evaluate the utility of MedicalNarratives, we train \model with a CLIP-like objective using our dataset spanning 12 medical domains.
\model outperforms previous state-of-the-art models on all 12 domains on a newly constructed medical imaging benchmark.
\href{https://huggingface.co/datasets/wisdomik/MedicalNarratives}{[Data]}

\end{abstract}    
\section{Introduction}
\label{sec:intro}


\begin{figure*}[ht!]
\centering
\includegraphics[width=\textwidth, height=0.55\textwidth]{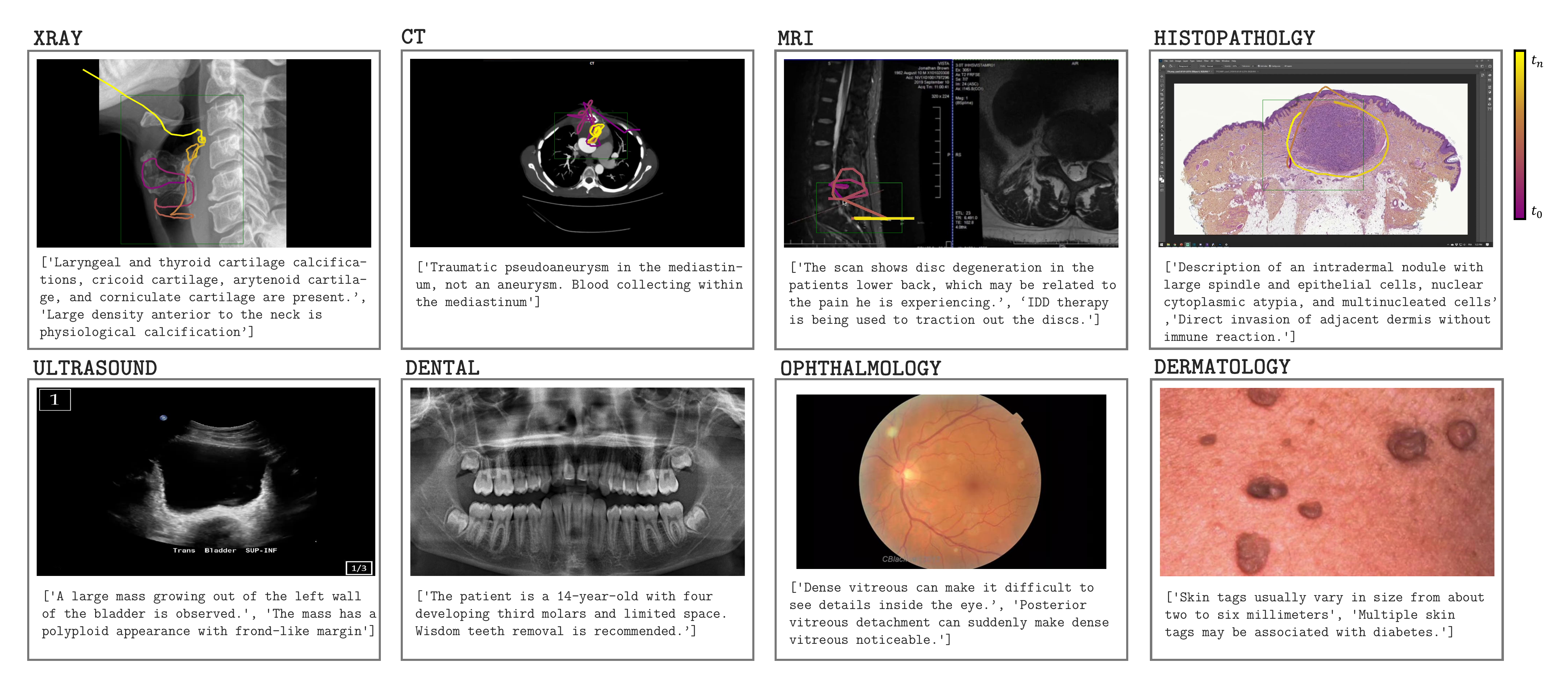}
   \caption{\textbf{\dataset}:Examples from our medical imaging modalities, excluding surgery, endoscopy, and general medical images due to their graphic nature. These samples are selected from interleaved video samples, with each sample showing the image, denoised text, and spatial traces \& bbox aligned in-time on 4 domains. See section \ref{supp:examples} in the Appendix for more examples and raw input text.}
\label{fig:main_data}
\end{figure*}

Analyzing medical images requires simultaneous spatial localization and semantic understanding \citet{perceptualconceptual}. An expert has to extract visual clues from image regions and combine them with retrieved knowledge from memory, arriving at a diagnosis. This process requires connecting individual spatial image regions to clinical concepts, often utilizing a segmental approach to avoid errors. \cite{perceptualconceptual}. In medical image analysis, typically semantic tasks like classification, captioning, and retrieval are explored exclusively from spatial tasks like detection~\cite{localizednar, xu2024pixellm, seyfioglu2024quilt}, or segmentation \cite{scribbleprompt, imis}. This can be attributed to the lack of large grounded multimodal datasets to train such models. Recent work like MedTrinity-25M \cite{medtrinity}, attempts to address this by releasing a multimodal dataset with spatial annotations, but relies on sub-optimally pretrained models to generate text descriptions and Regions of Interests (ROIs) for medical images lacking ground truth annotations, potentially propagating model biases and errors.

While data collection costs are steep, certain protocols balance ease of collection and training utility. Specifically, Localized Narratives \cite{localizednar, videolocalizednar} proposes a dataset of image, text, and grounding traces, curated by leveraging human annotators to describe an image vocally while simultaneously moving a computer mouse to the regions they describe, resulting in holistic grounded descriptions. This protocol of collecting grounded vision-language (VL) datasets does not have strict spatial annotations, yet, it captures strong spatial correlations to the description with every trace point, making the protocol uniquely easier to undertake and capture data en-mass as it appeals to the human nature to point while describing a scene \cite{kahneman1973attention, voigtlaender2023connecting, speakingpointing}. Localized narratives have been used to train models on semantic tasks \cite{localizednar, videolocalizednar, zhang2023pathnarratives}, and spatially-aware multimodal language models (MLM) like PixelLM \cite{xu2024pixellm}, and Molmo \cite{molmo} and other generative image models \cite{generateimagenar, partinar}.

To address these limitations, we draw inspiration from how medical experts naturally communicate and teach. In the joint field of cognitive psychology and medical imaging, studying how medical experts analyze patient data, studies leverage the think-aloud protocol \cite{durning2012using} to capture data for various types of analysis, where experts verbalize their thoughts as they perform a task, and some studies capture their eye gaze/cursor localizing the image regions they focus on \cite{thinkaloud, helle2017prospects}. This protocol has been used to collect medical datasets \cite{panetta2021tufts, molin2015slide}, including the Tufts dental x-ray database \cite{panetta2021tufts}, which captures a multimodal dataset incorporating radiologist expertise through eye-tracking and the think-aloud protocol.

We propose \dataset a dataset that leverages pedagogical medical videos where instructors narrate descriptions while pointing to relevant regions with their cursor, closely mimicking the think-aloud protocol used in clinical practice and the Localized Narratives protocol. Our dataset contains \countlen million image-text pairs across 11 medical modalities and 1 pseudo-medical domain, with interleaving samples between varying modalities (e.g., X-ray and CT for the same patient), which we argue improves downstream performance as these samples connect multiple visual and textual concepts. Importantly, 1M of these samples are grounded in expert traces that can be reformatted into bounding boxes or masks (see Figure \ref{fig:tr_to_seg}), serving as pretraining data for various tasks.


To test the base utility of our dataset, we train a vision-language model (\model) on our dataset and evaluate it on a new benchmark of datasets that cut across 11 medical modalities. On both classification and retrieval, we see our trained \model model outperform prior SOTA models like \biomed in both tasks with an average of 3\% and 14\% respectively. While the proposed dataset is a combination of data from \textbf{A.} Temporal unstructured sources like video, and \textbf{B.} Static structured sources like scientific articles, unlike prior work that solely leverage one source, our experiments show that the utility of the dataset increases with more data from video, with a net difference of 11.65\% on classification tasks and 53.6\% on zeroshot retrieval tasks. Finally, we show the utility of traces with qualitative examples, converting traces into segmentation using pretrained interactive medical image segmentation (IMIS) models like ScribblePrompt \cite{scribbleprompt, imis}. We hope future works leverage the dataset to train more grounded generative models similar to Quilt-LLaVA \cite{seyfioglu2024quilt}, LLaVA-Med++ \cite{medtrinity}, and PixelLM \cite{xu2024pixellm} as well as spatially-controlled medical image captioning models \cite{localizednar}. To bolster other use cases, we also release the constituting video clip IDs (useful for obtaining the videos) and many other metadata, including modality type and UMLS entities.

In addition to the centered still images with traces, we provide paired videos (temporal windows around trace start/end), preserving narration alignment to the pointing behavior. This addition allows models to learn spatiotemporal grounding (e.g., cursor trajectories across frames) rather than static spatial associations alone.


\begin{wrapfigure}{R}{0.45\textwidth}
    \centering
    \includegraphics[width=\linewidth]{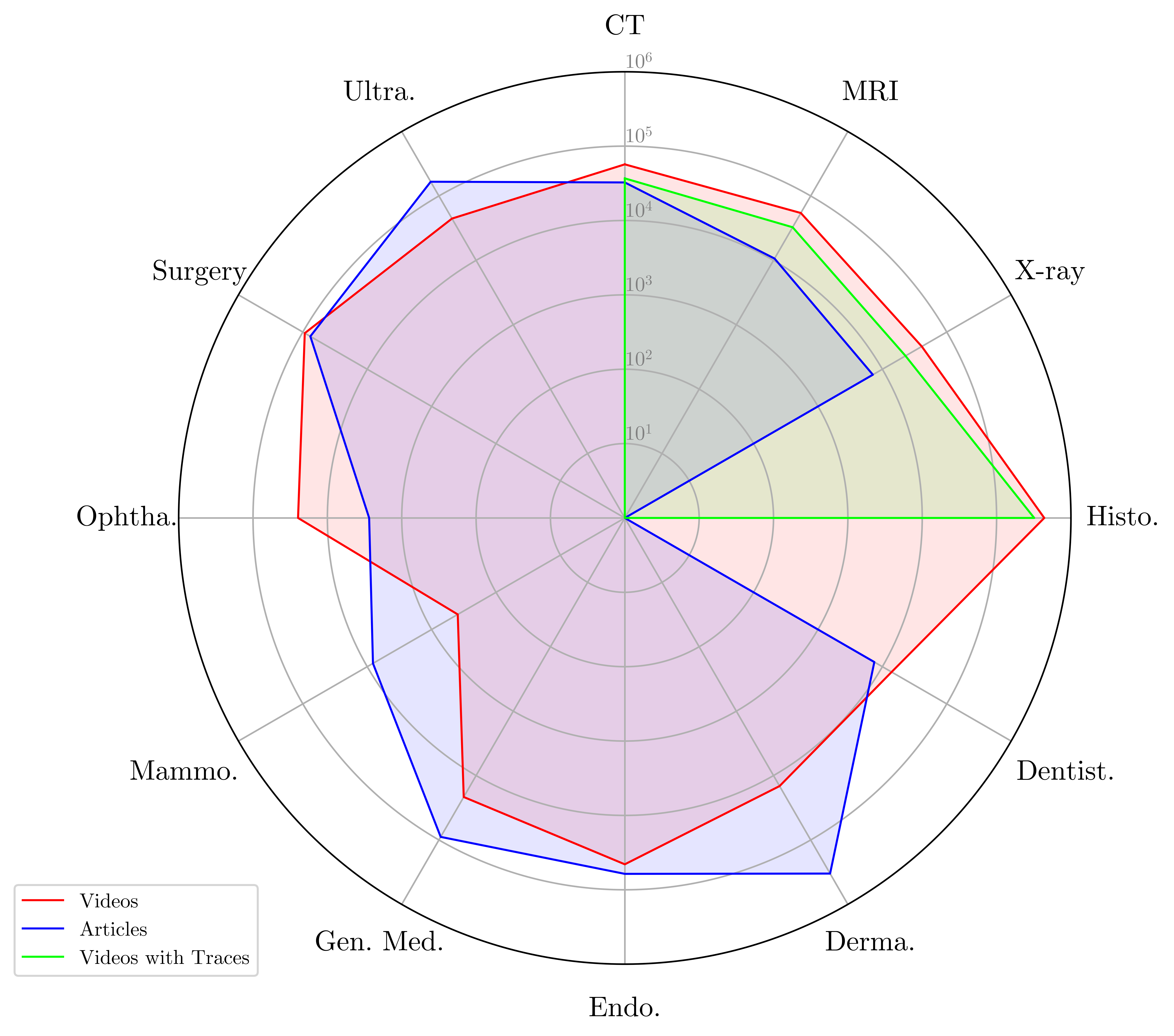}
    \caption{Breakdown of \dataset in size by modalities across both video and article subsets.}
    \label{fig:domain_breakdown}
\end{wrapfigure}



\begin{table*}[!b]
\centering
\scalebox{0.75}{
{\fontsize{8pt}{12pt}\selectfont
\begin{tabular}{l|c|c|c|c|c|c|c|c|c}
\toprule
\textbf{Dataset} & \textbf{Size}  & \textbf{Source} & \makecell{\textbf{Medical} \\ \textbf{Only}}  & \textbf{Domains} & \makecell{\textbf{Open Source} \\ \textbf{Data/Code}} & \makecell{\textbf{Novel}\\ \textbf{Images}} & \makecell{\textbf{ Video}} & \makecell{\textbf{ Text}} & \makecell{\textbf{Spatial}\\\textbf{Annot.}}\\
\midrule
\textbf{PMC-15M} \cite{biomedclip} & 15M & A & \xmark  & 30 & \xmark/\cmark & \cmark & \xmark  & Captions & \xmark \\
\textbf{PMC-FG-64M} \cite{biomedclip} & 46M & A & \xmark  & 30  & \xmark/\xmark & \cmark & \xmark  & Captions &  \xmark \\
\textbf{PMC-CLIP} \cite{lin2023pmc} & 15M & A & \xmark  & 12 & \cmark/\xmark & \cmark & \xmark  & Captions & \xmark \\
\textbf{MedTrinity-25M} \cite{medtrinity} & 25M & P  & \cmark & 10 & \cmark/\xmark & \xmark & \xmark  & Captions/labels & \makecell{Seg. mask} \\
\textbf{MedicalNarratives} & 4.7M & V+A  & \cmark  & 12 & \cmark/\cmark & \cmark & \cmark  & Expert & Traces \\
\bottomrule
\end{tabular}}}
\caption{\textbf{Comparison with large-scale medical datasets}. In the table, A: Articles, V: Videos, and P: pre-published datasets. Open-Source column is formatted \textit{data/pipeline}.}
\label{tab:dataset_comparison}
\end{table*}

\section{Related work}
\label{sec:related}

\subsection{Vision Language representation}
Vision-language (multi-modal) models have evolved over time in both supervised and self-supervised paradigms; in recent studies, contrastive self-supervision objectives \cite{radford2021learning, convirt, jia2021scaling} that learn by matching paired-modality embeddings have outperformed prior work \cite{vilbert, oscar, uniter} in downstream tasks and, more importantly, perform better at zeros-shot tasks or on emergent domains for which disparate modalities share a paired domain \cite{girdhar2023imagebind, zhu2023languagebind}. In medical imaging, early studies in radiology \cite{convirt, gloria} were pre-trained on specific x-ray images and their reports, and more recently domain specific VL models have pushed the SOTA on various tasks with models developed for Ophthalmology \cite{eyeclip}, Histopathology \cite{ikezogwo2024quilt, plip}, Computed Tomography \cite{ctclip}, Mammography \cite{mammoclip}, Dermatology \cite{monet}, Ultrasound/Echocardiography \cite{echoclip}. These models work well for the specific domains they are trained on and not for other domains, which may not have enough data to train for, hence the push for more general medical models \cite{biomedclip, zhou2024generalist, tu2024towards}.
 
\subsection{Medical (Localized) Narratives}
In training these VL models, much research effort is used in sourcing, filtering, and curating medical image(s)-text(s) paired data for pre-training, mostly sourcing general and specific medical domain datasets from Medical reports \cite{mimic}, PubMed \cite{pubmedclip, ruckert2024rocov2, arch, biomedclip}, books \cite{arch}, social-media \cite{plip, ikezogwo2024quilt}, YouTube/videos \cite{ikezogwo2024quilt, seyfioglu2024quilt} or mixtures of these \cite{medtrinity}. 

The utilization of these data for dense tasks like segmentation and detection (open-closed vocabulary) is limited as they do not provide any spatial annotation localizing regions of the images to specific labels/text, In contrast, every word in a localized narrative \cite{localizednar, videolocalizednar, zhang2023pathnarratives, seyfioglu2024quilt} is grounded to a region of the representative image by the point/trace captured. This datasets have been used to train models for semantic reasoning \cite{localizednar, videolocalizednar, zhang2023pathnarratives}, and for dense tasks \cite{panopticnarrative, openvocabwithnar, segnar}, and they also support training both generative multimodal language models \cite{xu2024pixellm, seyfioglu2024quilt, videolocalizednar} and generative image models \cite{generateimagenar, partinar}. Specifically in medical image analysis, Quilt-LLaVA \cite{seyfioglu2024quilt} adopts this paired data structure for training its histopathology chatbot with improved spatial reasoning, and Pathnarrative's \cite{zhang2023pathnarratives} hierarchical decision-to-reason localized narrative structure, enables classification and captioning tasks, offering explainable insights that improve human-AI collaboration in pathological diagnosis.

\section{\dataset:}
\label{sec:method}



Curating a vision-language dataset with spatial traces from unstructured pedagogy videos is a non-trivial, as many videos either lack voiced audio, fail to contain medically relevant content, or have insufficient medical relevance. In addition, conventional automatic speech recognition (ASR) systems also struggle with the specialized requirements of medical language transcription, necessitating a non-trivial solution. The de-noising of text and image modalities adds further complexity as the videos are typically conversational and, therefore, inherently noisy. Instructors often record both relevant and irrelevant visual content in their videos, making extracting frames at static intervals non-representative of the medical data contained in the video.

To collect \dataset, we leverage insights from Quilt-1M \cite{ikezogwo2024quilt} prior work, we trained models and handcrafted algorithms that leverage the nuances in the instructors' textual and visual behavior, ensuring accurate collection and alignment of both modalities. Finally, we manually filter noisy samples out and employ other heuristics and models to remove artifacts like faces and irrelevant traces. In this section, we start by characterizing the dataset \ref{xrizing_dataset}, then we discuss the methods used to source and filter the dataset \ref{sec:genquilt_main}, localize traces \ref{local_trace}, and discuss the implicit interleaving property \ref{interleave}. See Figure \ref{fig:video-pipeline} for our pipeline and section \ref{supp:A} and \ref{supp:B} of the appendix for how we align the data samples \ref{sec:alignment}.

\begin{figure*}[!ht]
    \centering
    \includegraphics[width=\textwidth]{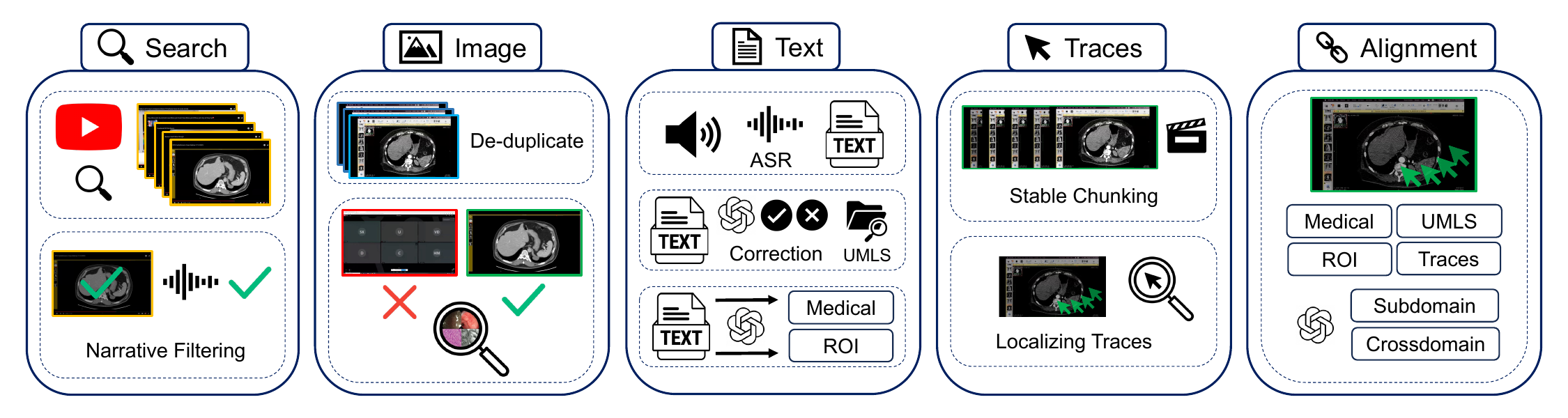} 
    \caption{The data curation pipeline for the Video subset of the \dataset dataset. \textbf{Search}: YouTube video-first search strategy, with filtering by pre-trained classifiers and heuristics. \textbf{Image}: Extracting keyframes of a video, denoising, and identifying medical images. \textbf{Text}: ASR transcription, text correction with LLMs, and medical/ROI text extraction. \textbf{Traces}: Identifying stable chunks of a video, then localizing cursor traces within each chunk. \textbf{Alignment}: Mapping medical/ROI text, traces, and images together. Samples are classified into finer-grained subdomains, and samples with discussions of multiple domains are identified with LLMs.}
    \label{fig:video-pipeline}
\end{figure*}

\subsection{Characterizing \dataset}
\label{xrizing_dataset}
To create \dataset we combine medical narratives curated from videos with image-text pairs curated from PubMed, resulting in 4.7M total image-text samples of which 1M samples are localized narratives. We compare our dataset against other medical pretraining datasets in Table \ref{tab:dataset_comparison} across various key distinctive properties including data source, and spatial annotation. Figure \ref{fig:domain_breakdown} shows the distribution of \dataset across various medical modalities, and Table \ref{roi-stats} and Table \ref{video-stats} in the appendix give detailed statistics across all medical modalities.

\subsubsection{Video-Subset}
We searched over 738K videos and extracted 74K narrative-styled videos that passed our heuristics and had relevant medical imaging pedagogy, a $10.1$\% yield making up a total of 4526 hours of video. In total, we collect 736K unique images with an average size of H: 1493px and W: 923px and 1.63M image-text pairs from videos with 1M of these samples grounded with traces, these samples cover 101.8M number of unique trace points yielding 547K number of unique bounding boxes with an average size of H: 316px and W: 357px across the 4 domains with traces: CT, MRI, X-ray, and Histopathology. The dataset contains 118K videos, collected at the boundaries of the traces, with a min, max, and average duration of 3.3, 228.8 and 24.2 seconds. The mean length of the text captions is 29.87 words, with an average of 2.48 medical sentences per image. Our dataset spans over 4M UMLS entities from those mentioned in the text with over 300K unique entities across medical (e.g., findings, or disease) and non-medical (e.g., governmental or regulatory activity) semantic types.

\subsubsection{Article Subset}
We extract 5.4M articles from PubMed \cite{pmc_open_access}, with 23M figures, after filtering for medical figures only, we obtain 1.03M figures from 273K articles, and after sub-figure separation, we have an average of 2.62 subfigure-subcaption pairs per-article figure, with an average of 45.45 words per-caption.



\subsubsection{Quality}
Unlike localized narratives \cite{localizednar, videolocalizednar} where localization accuracy can be measured by comparing against human annotation, none of our videos to our knowledge have any structured human spatial annotation to compare against. Nonetheless, to evaluate our pipeline’s performance, we assess several aspects. First, we calculate the precision of our LLM’s corrections by dividing the number of conditioned misspelled errors replaced (i.e., passed the UMLS check) by the total number of conditioned misspelled words found, yielding an average of 47.99\%. We also determined the unconditioned precision of the LLM, similar to the previous step, and found it to be 17.58\%. Therefore, we replace our detected incorrect words with the LLM’s correction 47.99\% of the time, and 17.58\% of the time we replace the LLM’s detected errors with its correction. To estimate the ASR model’s transcription performance, we compute the total number of errors replaced (both conditioned and unconditioned) and divide it by the total number of words in each video, resulting in an average ASR error rate of 0.81\%. Also note that, by prompting the LLM to extract only medically relevant text, we further eliminate identifiable information, such as clinic addresses, from our dataset.

Since the dataset was collected for pretraining, we do not upsample the text after correcting for errors and filtering bad images; on average, each image is paired with approx. 83 words of relevant text and traces when available and validated.


\subsection{Data Sourcing and Filtration} \label{sec:genquilt_main}
This involves (a) sourcing video/article data across 12 medical imaging domains, (b) filtering videos/articles, (c) denoising the captured images, captions, and trace modalities, and (d) aligning all modalities. We detail our method and highlight key contributions in sections \ref{supp:A} and \ref{supp:A} of the appendix.


\subsubsection{Text Extraction and Denoising}\label{text_denoising}
\textbf{Videos}: In line with Quilt-1M \cite{ikezogwo2024quilt} we leverage an open-source ASR model - Whisper \cite{radford2022robust} to transcribe all speech from the selected videos, correcting transcription errors using language model with specialized prompts (see section \ref{sup:text_denoising} for details on the error-extracting algorithm).

\noindent \textbf{Articles}: Similarly we parse each article's XML document, extracting each figure's caption and inline mentions (see \ref{sup:article_caption_extraction}). Since many sub-figures are typically grouped into single large figures, we split the compounded figure captions into sub-captions, leveraging an LM to find and split sub-captions due to the non-triviality of identifying enumerations in the text and splitting the captions correctly (see\ref{sup:subcap-sep}). Furthermore, we refine the inline mentions of a figure and match them to specific sub-captions/sub-figures (see \ref{sup:inlinepairing}).

\subsubsection{Image Extraction and Denoising} \label{img_denoise}

\textbf{Videos}: For each video, we identify medical key-frames and subsequently leverage these frames' times to split the video into time-intervals called {\it chunks} from which to extract representative image(s). To extract representative image(s), we use the median image of stable frames in each chunk if they exist, else we de-duplicate the captured key-frames, exploiting the human tendency in pedagogy videos to pause while explaining and pointing \cite{localizednar, speakingpointing, seyfioglu2024quilt}.

\noindent \textbf{Articles}: For scientific documents, we extract the figures as images. However, many of these figures contain multiple sub-figures which can take nonconventional grid shapes and are labeled irregularly, making the task of splitting into sub-figures and pairing with the correct sub-caption non-trivial. Since most compound figure layouts are not uniform and vary in the whitespace in between sub-figures, we train an object detection model based on the YOLO architecture \cite{yolov8} on sub-figure annotation datasets MedICaT and ImageCLEF 2016 \cite{medicat, imageclef2016}. See more details in section \ref{sup:subfig-detect}.

\subsection{Localizing Traces in Videos}\label{local_trace} Extracting the trace/cursor location from medical clips poses a significant challenge due to certain domain properties including homogeneity in color and texture, significant black/white background, and presence of artifacts in videos such as minor pixel variations and variations in the narrators' cursor movement speed and style. We modify the methodology proposed by Quilt-LLaVA \cite{seyfioglu2024quilt} centered around the observation that narrators typically pause before signaling with their cursor. We isolate segments in the video where the background is static, termed stable chunks. To do so, we develop a frame-differencing approach to detect chunks with minimal background movement. Our algorithm computes the absolute difference between consecutive frames and then applies a Gaussian filter for adaptive thresholding to detect frames with minor changes.


Due to the homogeneity of medical images, naive pixel-wise differencing produces many false positives, misidentifying changing chunks as stable. To mitigate this, we incorporate a perceptual metric, using the structural similarity index measure (SSIM) on randomly sampled patches to verify frame changes. Next, for each stable chunk, we compute a median frame (in the pixel domain) as a background reference, subtract it from individual frames, and apply a threshold to isolate the cursor. We then extract trace points by identifying the maximum pixel value coordinates. This method assumes minimal background motion, but subtle movements, such as narrator facial expressions, can interfere. To address this, we apply a face detection model \cite{schroff2015facenet} to mask distractions, ensuring focus remains on cursor movement. This algorithm provides a robust and generalizable approach for capturing cursor traces from medical videos.

\subsection*{Extracting videos} \label{video_extraction}
For each detected trace segment we extract the video clip at the start and end of the trace temporal window aligned with the text. 

\subsection{Cross-Modal Interleaved Samples} \label{interleave}

A key advantage of \dataset is its interleaved multi-modal nature. This manifests in two ways: \textbf{(1) Video-based interleaving:} Medical pedagogy videos frequently present multiple imaging modalities for the same patient. Instructors naturally explain relationships between these modalities in a single narrative, creating one-to-many mappings between textual descriptions and images. This allows our model to learn connections between modalities through shared textual context (see Figure ~\ref{fig:interleaved-examples} in Appendix). \textbf{(2) Sample-based interleaving:} Articles and Videos often contain images with multiple sub-images showing different modalities accompanied by a unified caption. This structure similarly reinforces cross-modal relationships. (see MRI example in Figure ~\ref{fig:main_data} and Figure ~\ref{fig:interleaved-examples} in the appendix). This interleaved nature of \dataset significantly enhances cross-modal retrieval capabilities, as shown in Sec. ~\ref{sec:crossmodal}. We open-source our dataset with modality tags which can be used to identify cross-modal samples.

\begin{figure*}[h!]
\centering
\includegraphics[width=\textwidth]{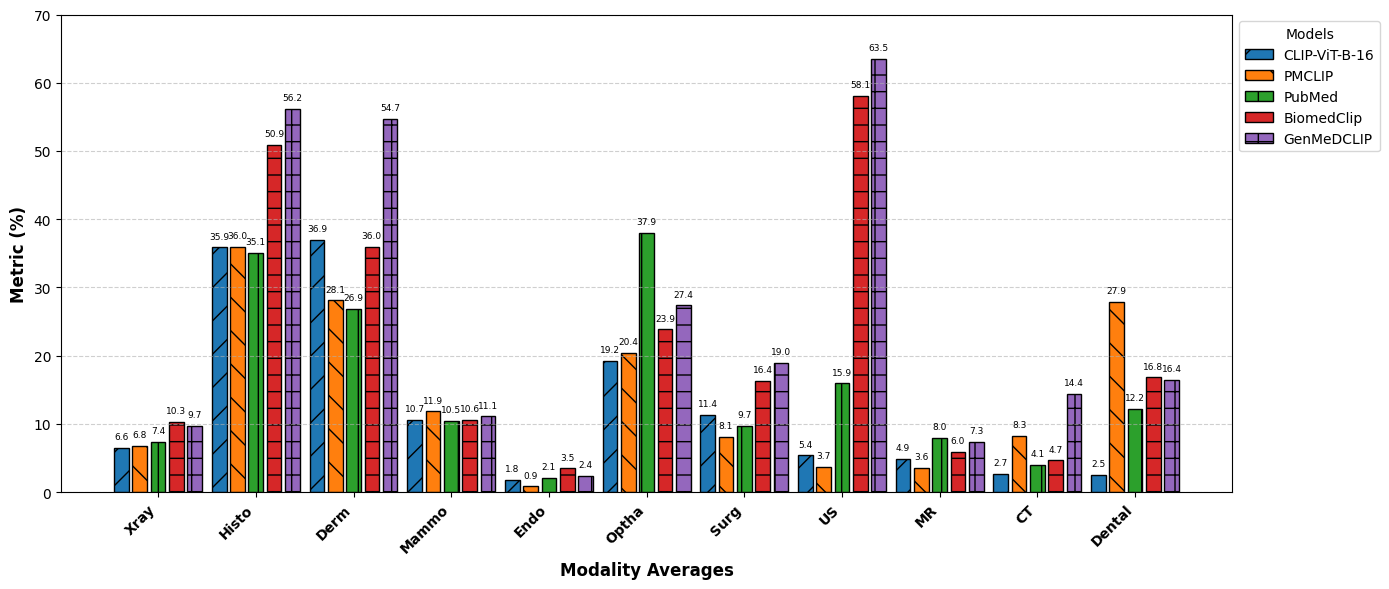}
   \caption{\textbf{Zeroshot Classification Results} shows that our model \model outperforms all other baselines including the out-of-domain CLIP, and biomedical vision-language models \biomed, and \pubmedclip across the constructed medical benchmark which covers all 11 medical domains represented. The metric for Xray and Mammography is mean average precision while the rest is accuracy.}
\label{fig:zeroshot_clf}
\end{figure*}


\section{\model: Experiments}
\label{sec:training}

We test the utility of \dataset on two medically relevant tasks image classification (zeroshot and linear probing) and cross-modal information retrieval (zero-shot) across all in-domain modalities. We select the Contrastive Language-Image Pre-training (CLIP) objective \cite{radford2021learning} to pre-train a VL model: \model. We train several models, varying the image, and text encoders while making adaptations in line with prior work on the choice of encoders and text tokenization for improved performance \cite{biomedclip, ikezogwo2024quilt}. For the image tower, we finetune Vision Transformers (ViT-Base) \cite{dosovitskiy2020vit} models pretrained using a supervised cross-entropy objective (ViT-Base-16 and ViT-Base-32 \cite{rw2019timm}) and unsupervised contrastive objective ViT-Base-16) \cite{radford2021learning}, on 224*224 pixel images. On the text tower, we use GPT2 \cite{radford2019language} with a context length of 77, and BioMedBert \cite{biomedbert} with context size to 256. To train our models we utilize OpenClip \cite{ilharco_gabriel} on 4 Nvidia A40 GPUs for 20 iterations. To ensure a fair comparison with baselines, we trained three different variants of our model: \modelthreetwo: with ViT-B/32 image-tower and GPT2/77 text-tower architecture, \modelPMB: with ViT-B/16 image-tower and Bert/256 BiomedBert \cite{biomedbert} text-tower, and \modelPMB: with ViT-B/16 image-tower and GPT2/77 \cite{biomedbert} text-tower; all finetuned for 20 epochs over our train-set, while data split ablation models are trained for 6 epochs. (Details in Section \ref{sup:training_details} in Appendix.)

\begin{figure}[!ht]
    \centering
    \includegraphics[width=0.75\linewidth]{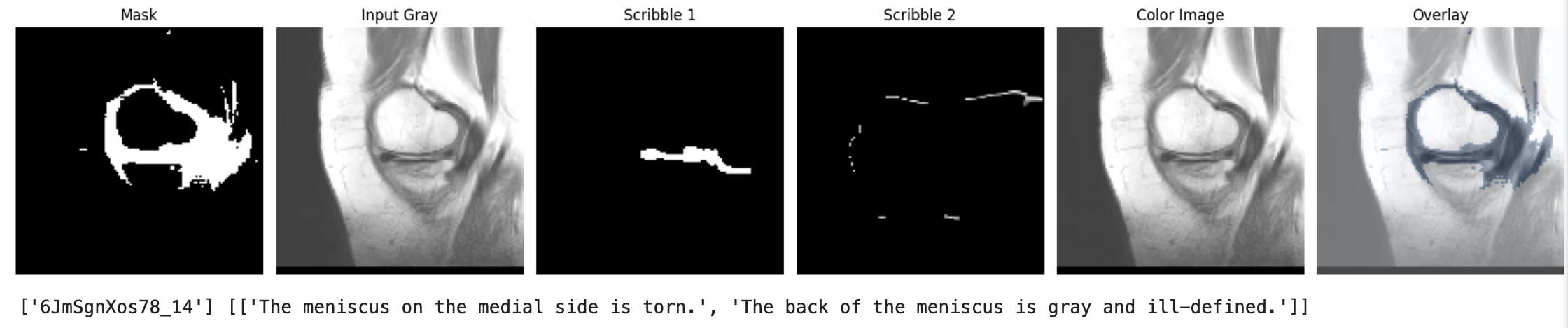}\\[3pt]
    \includegraphics[width=0.75\linewidth]{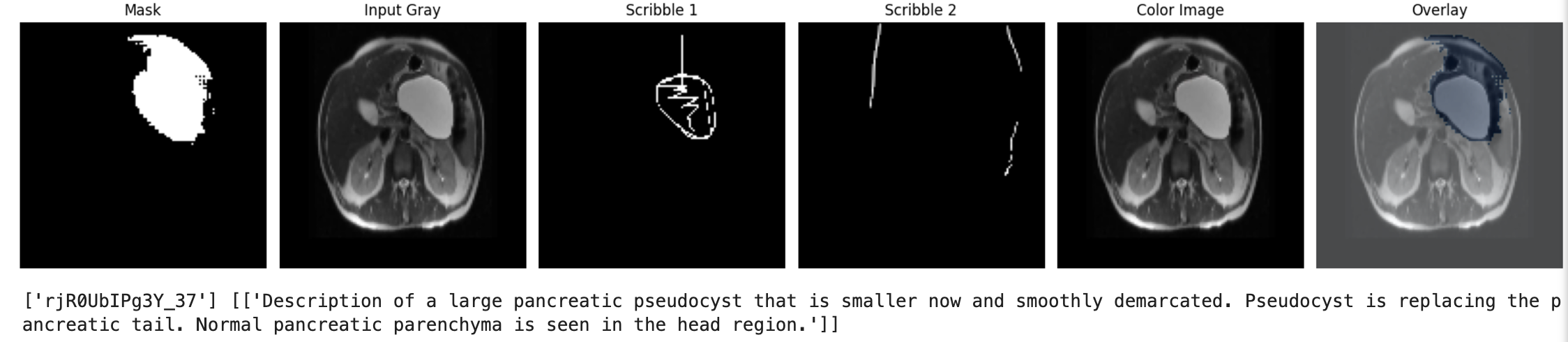}\\[3pt]
    \includegraphics[width=0.75\linewidth]{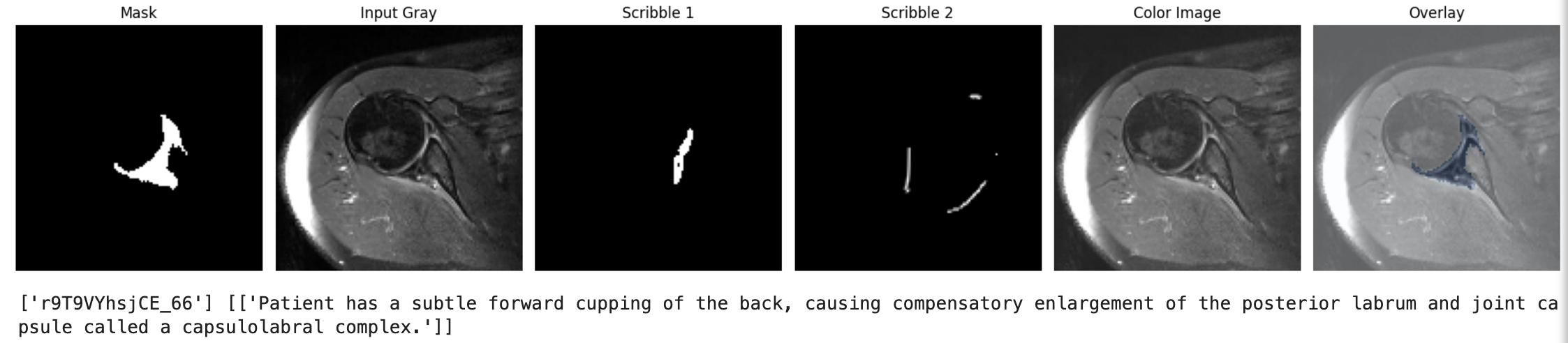}
    \caption{\textbf{Using trace as prompts for segmentation using ScribblePrompt-SAM}. \textit{(Right)} resulting mask from trace \textit{(Center)}.}
    \label{fig:tr_to_seg}
\end{figure}

\setlength{\tabcolsep}{1pt}
\begin{table}[h!]
\centering
\tablestyle{3pt}{1.2}
\begin{tabular}{l|cccccc|c}
\hline
\textbf{Model} & \textbf{Isic} & \textbf{Til} & \textbf{Pcam} & \textbf{Mhist} & \textbf{Nck} & \textbf{Mammo} & \textbf{Avg} \\
\toprule

CLIP-ViT-B-16 \cite{radford2021learning} & 71.23 & 91.23 & 82.42 & 63.97 & 92.26 & 83.30 & 80.74 \\
\pubmedclip \cite{pubmedclip} & 68.58 & 91.32 & 84.07 & \textbf{72.16} & 92.29 & 83.90 & 82.06 \\
\biomed \cite{biomedclip} & 68.25 & 91.82 & 83.43 & 66.73 & \textbf{93.05} & 83.70 & 81.17 \\
\hline
\modelthreetwo & 72.75 & 93.26 & 86.77 & 72.06 & 92.77 & 83.70 & 83.55 \\
\modelPMB & 69.38 & 91.51 & 84.54 & 67.66 & 88.02 & 84.20 & 80.88 \\
\model & \textbf{74.87} & \textbf{93.34} & \textbf{87.69} & \textbf{72.16} & 90.84 & \textbf{84.90} & \textbf{83.97} \\
\bottomrule
\end{tabular}
\caption{\textbf{Linear Probing} results across datasets representing Dermatology (Isic), Histopathology (pcam, mhist, nck), and Mammography (vinDr-Mammo) classification tasks. \model outperforms all baselines showing the capacity of our model to be fine-tuned for downstream tasks. The metric used is accuracy.}
\end{table}

\subsection{Benchmarking on Downstream Medical Tasks}
 We evaluate the utility of \model on a new medical imaging benchmark of all medical domains represented in our pre-training dataset \dataset, with some domains represented by $>=1$ dataset/task for classification, totaling 29 downstream datasets and on a held-out set of 1000 unique images for the retrieval task downstream.
For MRI, CT, and ultrasound we use their respective subsets from \textbf{RadImageNet} \cite{mei2022radimagenet} dataset. For Xray, we evaluate on \textbf{VinDr-CXR} Chest Xrays \cite{nguyen2022vindr} test set and report the mean average precision (mAP), similarly for Mammography we use \textbf{VinDr-Mammo} \cite{vindrmammonguyen} and report the mAP. We evaluate surgical organ classification using \textbf{Dresden} \cite{carstens2023dresden}, and for endoscopy, we test on all procedure images in \textbf{GastroVison} \cite{gastrovision}. For Dermatology we evaluate on the \textbf{Diverse Dermatology Images} (DDI) \cite{ddi} binary (benign or malignant) dataset and Isic 2018 dataset \cite{isic1}.
For Dentistry we evaluate on \textbf{Dental orthopantomography} (OPG) \cite{rahman2024dental} X-ray dataset. To evaluate the Ophthalmology domain we evaluate on \textbf{G1020} \cite{bajwa2020g1020} a retinal fundus glaucoma dataset and on \textbf{Optical Coherence Tomography Dataset} (OCTDL) \cite{octdl}. We evaluate the Histopathology domain on the following datasets: \textbf{PatchCamelyon} \cite{veeling2018rotation}, \textbf{NCT-CRC-HE-100K} \cite{kather2018100}, \textbf{BACH} \cite{aresta2019bach}, \textbf{Osteo} \cite{arunachalam2019viable}, \textbf{SkinCancer} \cite{kriegsmann2022deep}, \textbf{MHIST} \cite{wei2021petri}, \textbf{LC25000} \cite{borkowski2019lung}, and on TCGA-TIL \cite{tcga-til}. Please see section \ref{sup:training_details} in the appendix for more details.

\setlength{\tabcolsep}{1pt}
\begin{table}[ht!]  
\centering 
\tablestyle{3pt}{1.2}
\begin{tabular}{l|c|ccc|ccc|c}  
\toprule
    \textbf{Models} & \textbf{Data}  & \multicolumn{3}{c|}{\bf T2I retrieval} & \multicolumn{3}{c|}{\bf I2T retrieval} & {\bf Avg}  \\
    &  & @5  &  @50  &  @200  &  @5  &  @50  &  @200  &    \\
    \shline
CLIP-ViT-B-16 \cite{radford2021learning} & - & 3.48 & 20.38 & 35.69 & 3.56 & 20.39 & 35.42 & 19.82 \\
PMC-CLIP \cite{lin2023pmc} & A & 0.01 & 0.33 & 1.18 & 0.01 & 0.34 & 1.24 & 0.52 \\
\pubmedclip \cite{pubmedclip} & A & 1.44 & 12.68 & 25.44 & 1.10 & 12.30 & 24.07 & 12.84 \\
\biomed \cite{biomedclip} & A & 16.50 & 51.48 & 67.46 & 15.71 & 48.85 & 64.61 & 44.10 \\
\hline
\modelthreetwo & V+A & 22.36 & 76.33 & 88.60 & 20.75 & 75.15 & 88.23 & 61.90 \\
\modelPMB & V+A & 28.29 & 82.91 & \textbf{92.43} & 29.21 & 82.91 & \textbf{92.43} & 68.03 \\
\model  & V+A & \textbf{34.89} & \textbf{83.83} & 92.27 & \textbf{34.26} & \textbf{83.48} & 92.32 & \textbf{70.17} \\
\hline \multicolumn{9}{c}{\textbf{ Data Split Ablation}} \\
\hline
\model$^*$ & A & 2.11 & 12.89 & 22.36 & 2.35 & 13.66 & 22.81 & 12.70 \\
\model$^*$ &  V+A & 28.01 & 80.56 & 90.96 & 27.48 & 79.95 & 90.85 & 66.30 \\
\bottomrule
\end{tabular}  
\caption{\textbf{Retrieval} results on our held-out set of 16K samples across all medical domains, show that our model \model outperforms all other baselines on both Zeroshot image-to-text and vice-versa text-to-image retrieval task. In the table, A: Articles, V: Videos, and $^*$ represents a shorter number of fine-tuning iterations}
\label{tab:retrieval}  
\end{table}

\subsection{Zero-shot classification }
We evaluate our model’s zero-shot performance against three state-of-the-art models: CLIP, \biomed, \pmcclip, and \pubmedclip. In Figure \ref{fig:zeroshot_clf} and Table \ref{tab:specificzeroshotclf}, each domain in the benchmark is represented by a set of dataset(s). The prompts used for these evaluations are presented in Table \ref{tab:text_prompts_clf} in the Appendix.  Across the benchmark, our model averages the following \modelthreetwo: 31.33\%, \modelPMB: 31.46\%, and \model: 32.55\% metric all outperforming \biomed with 27.80\% by 4.75\%. Specifically, as shown in Figure \ref{fig:zeroshot_clf}, \model outperforms all baselines in five medical domains: Histopathology, Dermatology, Surgery, Ultrasound, and CT, while remaining comparable to baselines in the Chest X-ray, Endoscopy, Mammography, and MRI domains.

\subsection{Supervised linear probing}
We assess the full-shot performance of our model by conducting linear probing with 100\% of the training data; we report the average accuracy over all benchmark evaluation across five distinct datasets, specifically those with dedicated training and testing sets among our external datasets in Dermatology, Histopathology, and Mammography. Remarkably, our model, utilizing the ViT-B/32 architecture with GPT/77, outperforms its counterparts, \biomed, and CLIP, in most datasets. Overall, on average \model outperforms all other models including \biomed and \pubmedclip with over 2.8\%, and over 1.9\% respectively.

\subsection{Cross-Modal Retrieval }
\label{sec:crossmodal}
We evaluate cross-modal retrieval performance by examining both zero-shot text-to-image and image-to-text retrieval capabilities. To do so, we leverage a randomly selected held-out partition of \dataset, not used in training our models. The held-out set contains 16K image-text pairs with the following medical modality distribution: 1756 X-ray, 1237 MRI, 1851 CT, 1351 Ultrasound, 1744 Surgery, 1346 Endoscopy, 1189 Dermatology, 1216 Dentistry, 1151 Ophthalmology, 1000 Histopathology, 1299 General Medical, 1149 Other (Mammo etc) image-text pairs. Retrieval in our study is done by identifying the nearest neighbors for each modality and then determining whether the corresponding pair is within the top N nearest neighbors, where N $\in$ {1, 50, 200}, mimicking several medical search tasks. Results in Table \ref{tab:retrieval} shows that on average \model outperforms all baselines and specifically outperforms \biomed by 26.07\%, The results also confirm the observation made in \biomed \cite{biomedclip} where the general CLIP model outperforms the in-domain model \pubmedclip by 6.98\%

\subsection{Data Split Ablation}
As seen in Tables \ref{tab:specificzeroshotclf} and \ref{tab:retrieval}, we ablate the added utility of capturing pedagogy video data by training two models, one trained solely on articles and the other on both articles and video data. The results show that adding the video-derived data leads to higher average classification (11.65\% higher) and retrieval (53.6\% higher) performance. We suspect that the dynamic nature of \textit{video frames} introduces diverse vantage points, partially explaining these improvements. We also see that classification performance across all Article only trained models except Biomedclip is comparable further buttressing the impact of video as a data source.



\section{Discussion and Limitations}
\label{limitations}
\dataset contributes a robust pipeline for grounded multi-modal data curation across noisy, unstructured, and diverse medical modalities sources. We believe it would catalyze progress in novel medical vision-language tasks, like spatially-controllable report generation \cite{localizednar, xu2024pixellm}, and interactive medical image segmentation \cite{imis, scribbleprompt}. Figure ~\ref{fig:tr_to_seg} illustrates how the captured traces, albeit noisy and not expert-validated, can serve as conditioning for semi-automatic segmentation models like ScribblePrompt \cite{scribbleprompt} toward plausible object boundaries and for exploring visual grounding toward text+trace conditioned segmentation.

\subsection*{Spatial Reasoning Applications}
Beyond retrieval/classification, the trace-aligned samples provide direct supervision for grounded language and localization tasks without dense masks. Following Localized Narratives \cite{localizednar, videolocalizednar}, each word/phrase co-occurring with a trace segment supplies weak phrase-region links for grounded captioning, referring expressions, and spatial relation inference. High-dwell (i.e. spatial regions where narrators focus on) segments of traces can be collapsed into pointing cues to train pointing-based medical MLMs, such as Molmo \cite{molmo}. Because our traces are timestamped, the same supervision naturally extends to video: the trajectory of the cursor across frames yields spatiotemporal grounding suitable for dynamic "point-while-describe" models and temporal localization (e.g., axial CT sweeps or ultrasound). The dataset also supports tasks that predict spatial traces as additional loss objectives toward imbuing models with spatial understanding \cite{xu2024pixellm} and panoptic narrative grounding objectives \cite{panopticnarrative} are directly enabled by these trace–text alignments.

For dense prediction and controllable generation, traces act as sparse supervision that can be transformed into inputs for interactive medical image segmentation (IMIS) models (e.g., as points/scribbles for ScribblePrompt) to bootstrap pseudo-masks and iteratively refine them \cite{scribbleprompt}. Coupling trace spans with their co-mentioned phrases supplies approximately localized phrase labels for open-vocabulary detection/segmentation similar to phrase-region training \cite{segnar, openvocabwithnar}.

Finally, the same signals can guide where and what to synthesize in text-to-image/volume models, using trace/point conditioning alongside clinical text to localize clinical entities, while enabling spatially controllable medical report generation \cite{generateimagenar, partinar, seyfioglu2024quilt, xu2024pixellm}.

\subsection*{Limitations}
\begin{enumerate}
    \item Our dataset lacks human-annotated bounding boxes, limiting overlap assessment between video traces and annotations, restricting dense tasks like semantic segmentation.
    
    \item Our dataset overrepresents abnormal cases, an underlying bias, reflected in hospital practices where imaging follows clinical suspicion. This may impact model generalizability and introduce bias in clinical decision support.

    \item While we showcase the capacity of traces to be useful for IMIS task, this work does not leverage the traces to train any models toward downstream spatial or spatial aware models like PixelLM \cite{xu2024pixellm}. We leave this to future work.

    
\end{enumerate}

\section{Conclusion}
\label{sec:conclusion}

This study proposes a robust protocol for collecting and annotating medical narratives. Our curated dataset \dataset, which follows the Narratives Annotation Protocol addresses the specific challenges of medical data collection at scale balancing the relationship between downstream utility and ease/cost of collection. We argue that this protocol can serve as the de facto standard for annotating future multimodal medical datasets, particularly given its flexibility in capturing grounded text describing medical images effectively. We demonstrate a strong performance over prior models, across both classification and retrieval tasks, establishing new state-of-the-art results and demonstrating the effectiveness of data filtration methods on model performance, as we train our \model on \countlen samples while \biomed trains on over 15M samples. We hope future work leverages our developed models, dataset, and protocol.


\section{Acknowledgments}
\label{sec:awk}
We thank Microsoft for providing OpenAI credits. This work is also funded through a Population
Health Initiative at the University of Washington.







\bibliographystyle{abbrvnat}
\bibliography{main}  


\newpage

\appendix

\section{\dataset: Video Curation}
\label{supp:A}


\begin{figure*}[!ht]
    \centering
    \includegraphics[width=\textwidth]{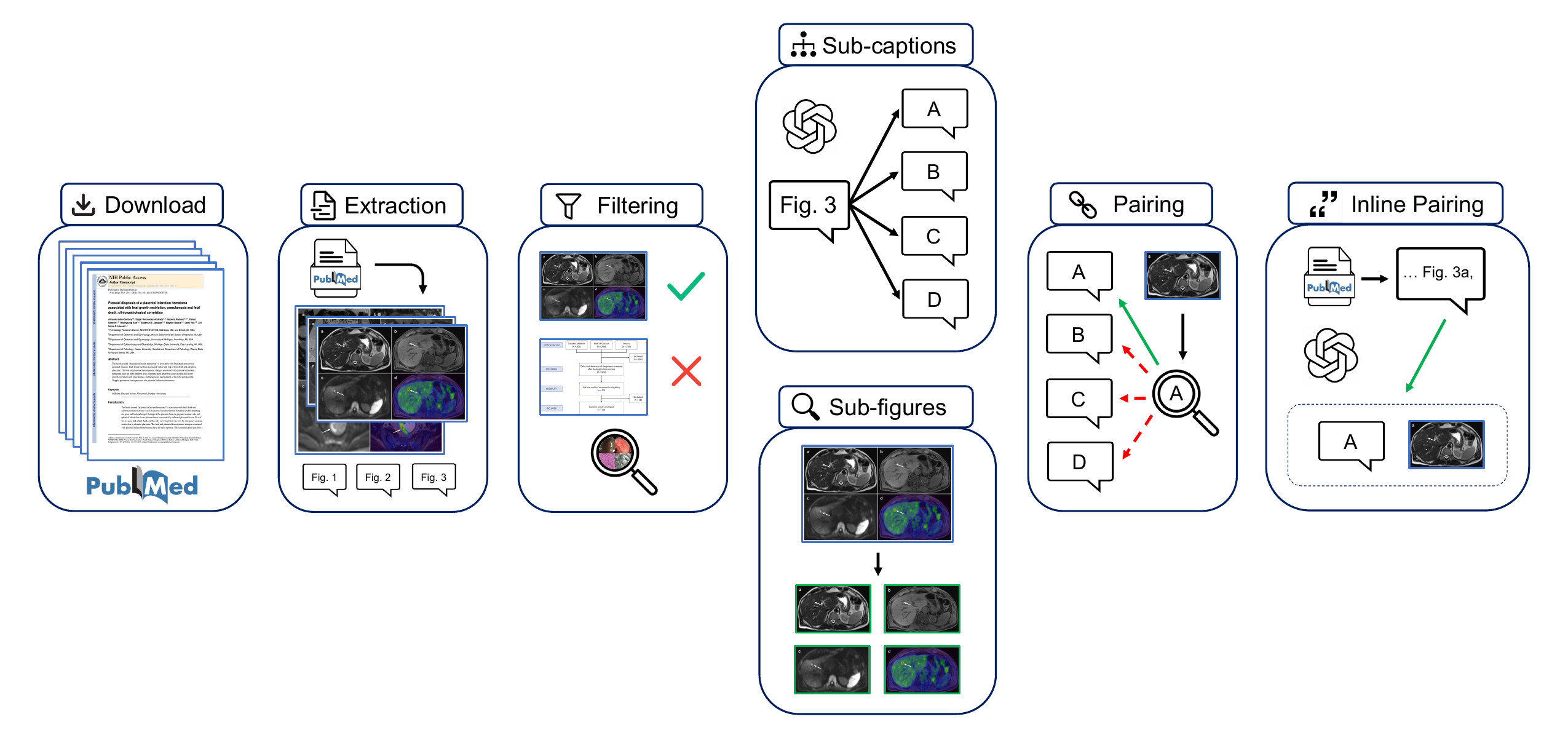} 
    \caption{The data curation pipeline for the PubMed subset of the \dataset dataset. \textbf{Download}: downloading PMC-OA. \textbf{Extraction}: extracting figures, captions, and inline references. \textbf{Filtering}: filtering for medical images. \textbf{Sub-captions}: splitting compound figure captions into sub-captions. \textbf{Sub-figures}: detecting and cropping sub-figures from compound figures. \textbf{Pairing}: matching sub-figures and sub-captions. \textbf{Inline pairing}: matching inline mentions of figures with the most relevant sub-figure or sub-figures.}
    \label{fig:pubmed-pipeline}
\end{figure*}

Distilling the volume of data YouTube offers into a grounded vision-language dataset that captures the all available signal of medical pedagogy video data is a significant task. Each step in the data curation process presents unique challenges when scaling to handle multiple medical domains.

With \dataset, we collect vision-language datasets grounded in time with language-correlated traces across twelve medical domains with the first three domains defined to be \textit{static} where representative samples are usually static images: (1) computed tomography (CT), (2) magnetic resonance imaging (MRI), and (3) xray, and \textit{non-static} domain with representative samples exhibiting significant visual change: (4) ultrasound, (5) surgery, (6) endoscopy, (7) dentistry, (8) dermatology, (9) mammography, (10) ophthalmology, (11) histopathology and (12) general medical illustrations. When processing these subsets, our approach differs to accommodate the nuances of the video data. Our data curation pipeline can be split into these high-level tasks of:

\begin{enumerate}[label=(\Alph*)]
\item Searching for representative videos in each medical domain.
\item Filtering videos for narrative style.
\item Extracting image, text, and cursor traces from selected videos.
\item Denoising and de-duplicating the collected raw data.
\item Aligning image, text, and localization traces.
\item Collecting metadata useful for varying downstream tasks (e.g. subdomains) and interleaving the dataset.
\end{enumerate}

\begin{figure*}[ht!]
    \centering
    \includegraphics[width=\textwidth]{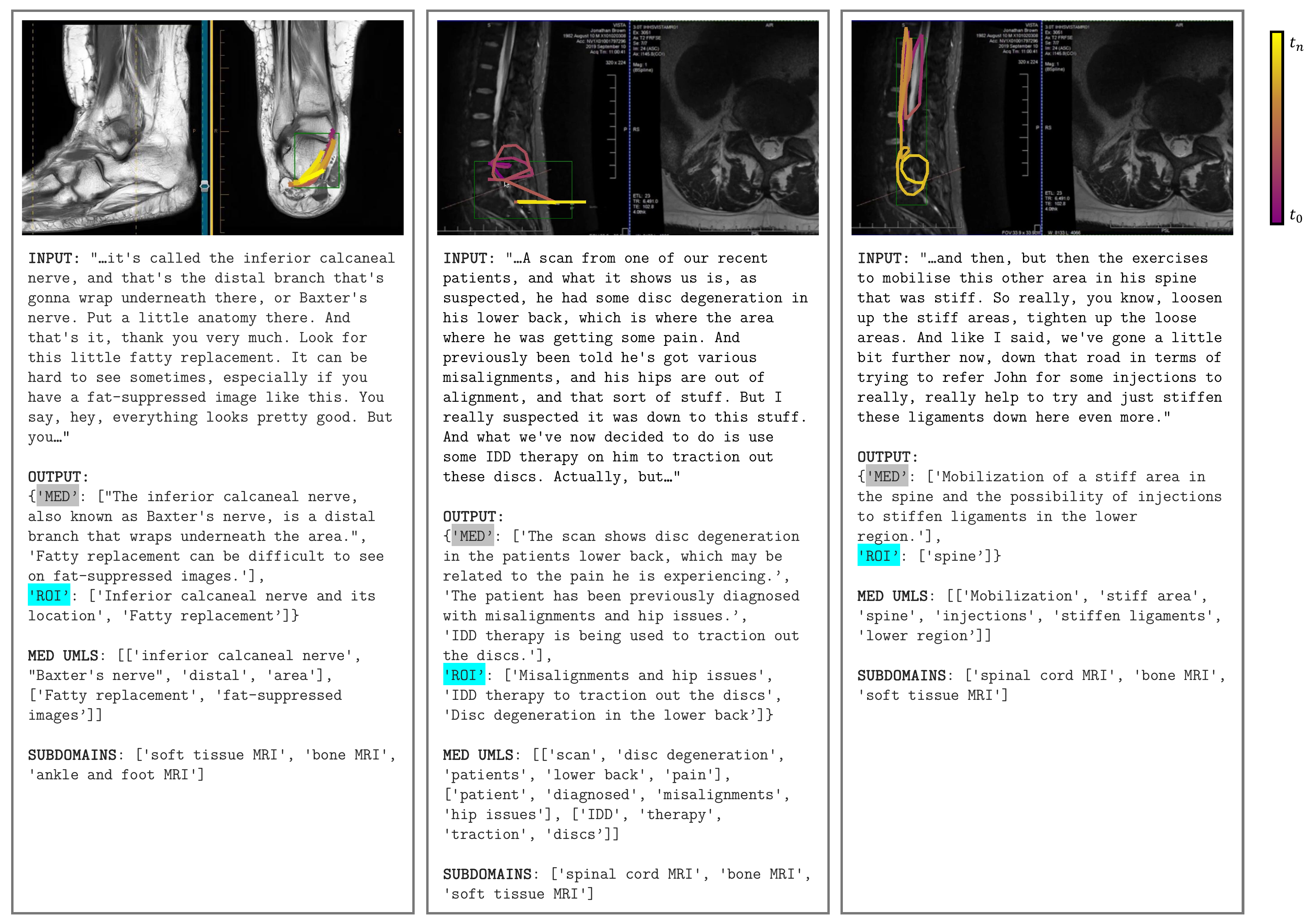} 
    \caption{\textbf{MRI examples} with in the \dataset dataset. \textbf{Input}: raw input text from ASR. \textbf{Output}: the output from the LLM, with denoised medical and ROI text. \textbf{Traces}: Cursor traces and bounding boxes aligned in-time with the raw text. \textbf{UMLS}: UMLS entities extracted from the medical text. \textbf{Subdomain}: classification of the sample into finer-grained subdomains.}
    \label{fig:mri-roi-examples}
\end{figure*}

\begin{figure*}[ht!]
    \centering
    \includegraphics[width=\textwidth]{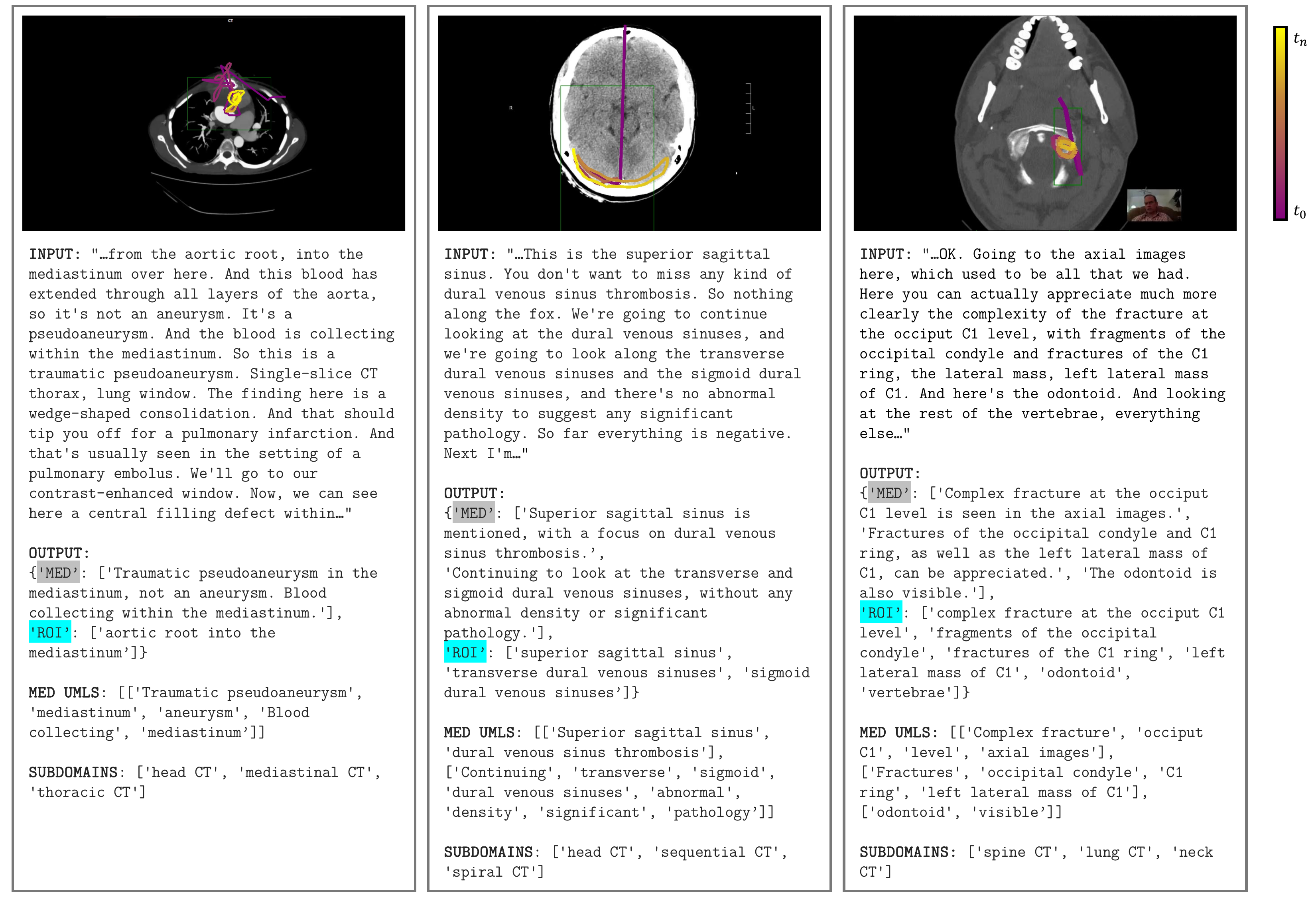} 
    \caption{\textbf{CT examples} with in the \dataset dataset. \textbf{Input}: raw input text from ASR. \textbf{Output}: the output from the LLM, with denoised medical and ROI text. \textbf{Traces}: Cursor traces and bounding boxes aligned in-time with the raw text. \textbf{UMLS}: UMLS entities extracted from the medical text. \textbf{Subdomain}: classification of the sample into finer-grained subdomains.}
    \label{fig:ct-roi-examples}
\end{figure*}

\begin{figure*}[ht!]
    \centering
    \includegraphics[width=\textwidth]{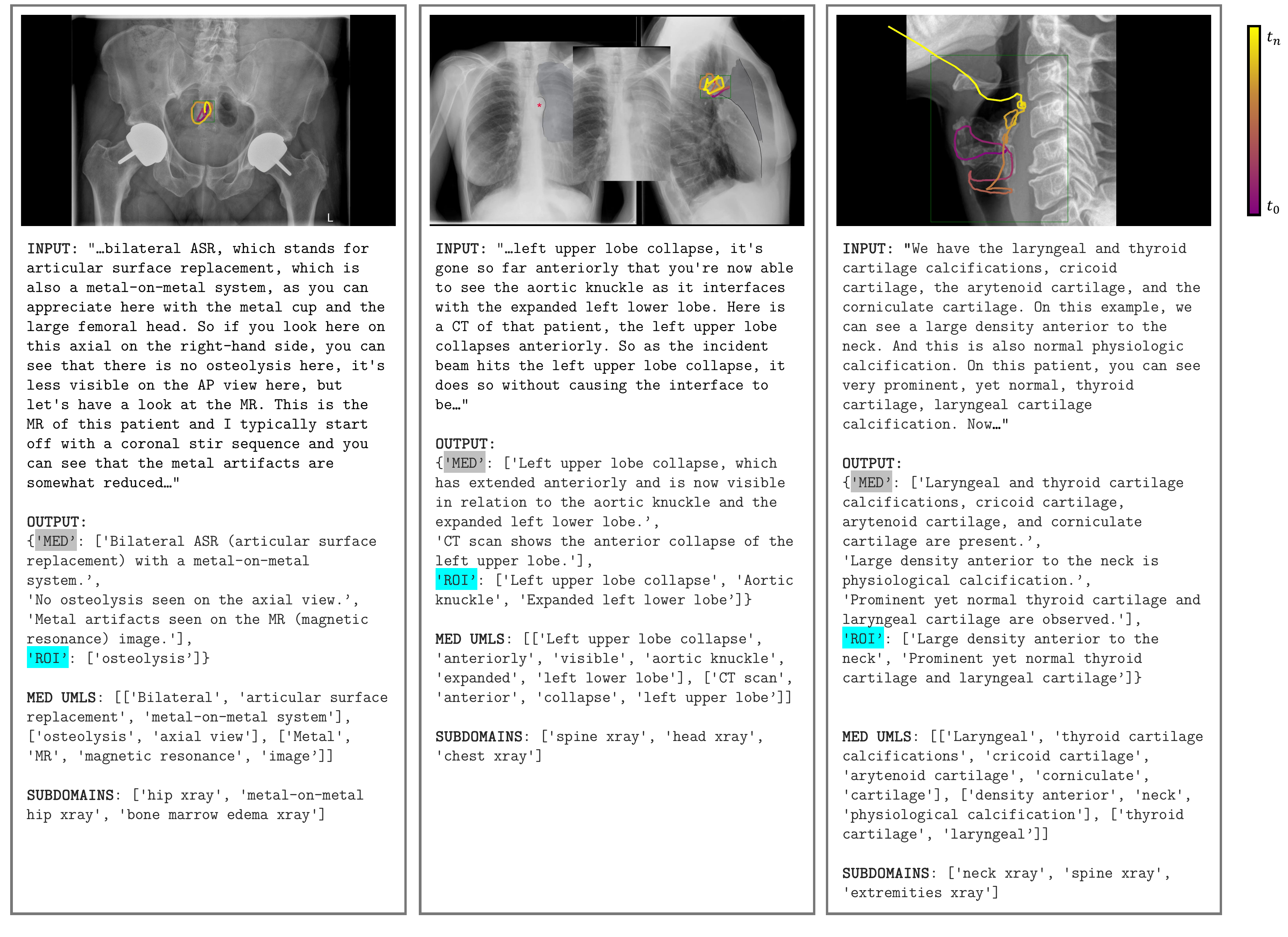} 
    \caption{\textbf{X-ray examples} within the \dataset dataset. \textbf{Input}: raw input text from ASR. \textbf{Output}: the output from the LLM, with denoised medical and ROI text. \textbf{Traces}: Cursor traces and bounding boxes aligned in-time with the raw text. \textbf{UMLS}: UMLS entities extracted from the medical text. \textbf{Subdomain}: classification of the sample into finer-grained subdomains.}
    \label{fig:xray-roi-examples}
\end{figure*}

\begin{figure*}[ht!]
    \centering
    \includegraphics[width=\textwidth]{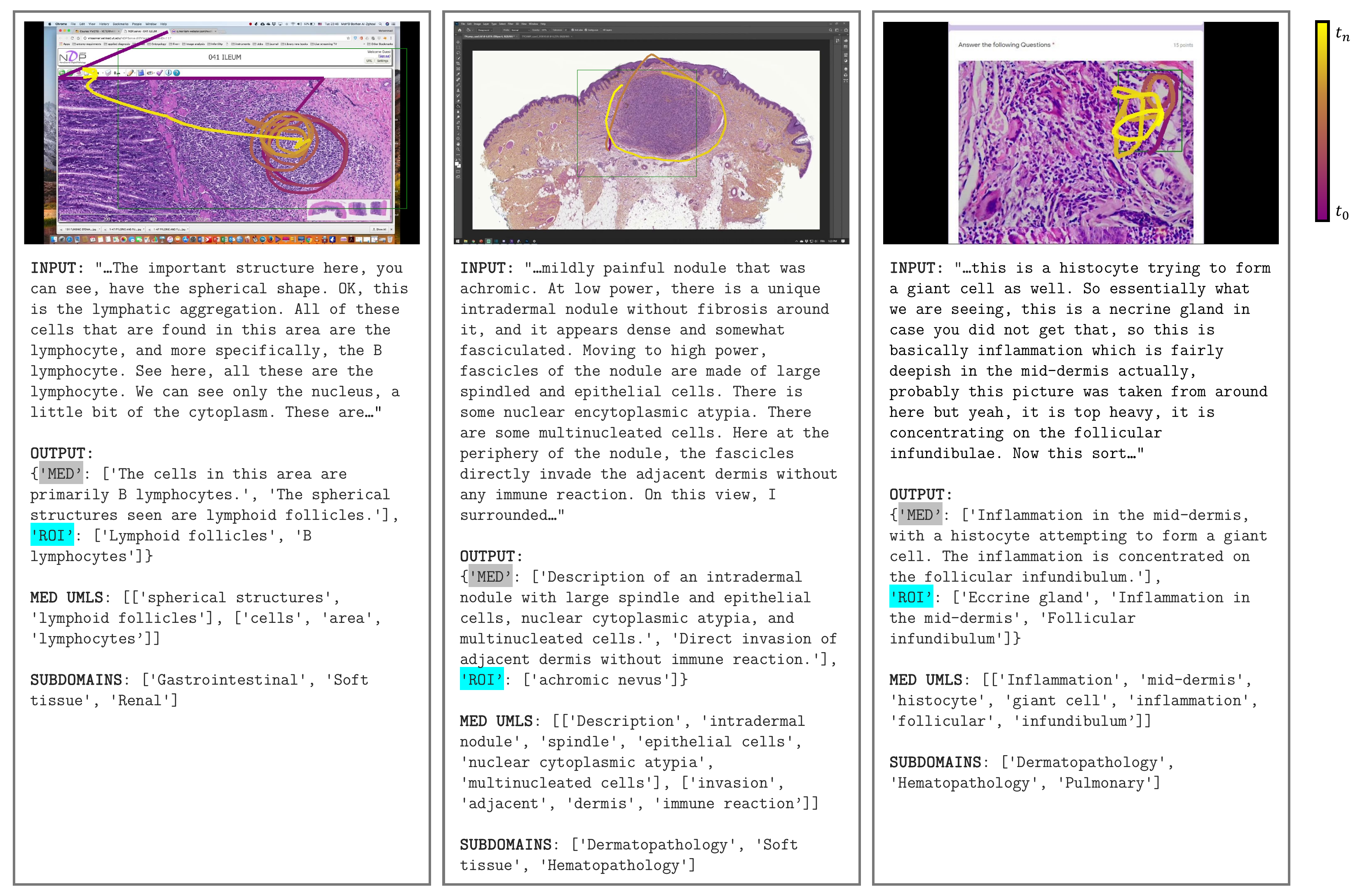} 
    \caption{\textbf{ Histopathology examples} within the \dataset dataset. \textbf{Input}: raw input text from ASR. \textbf{Output}: the output from the LLM, with denoised medical and ROI text. \textbf{Traces}: Cursor traces and bounding boxes aligned in-time with the raw text. \textbf{UMLS}: UMLS entities extracted from the medical text. \textbf{Subdomain}: classification of the sample into finer-grained subdomains.}
    \label{fig:histo-roi-examples}
\end{figure*}

In the following sections, we present a detailed overview of the major steps in curating \dataset starting with search. We also present examples of curated narrative samples in Figures \ref{fig:xray-roi-examples}, \ref{fig:histo-roi-examples}, \ref{fig:mri-roi-examples}, and \ref{fig:ct-roi-examples}.

\subsection{Domain-Specific Search}\label{sec:domain_specific_search}

We first identify medical channels and videos for each domain on YouTube, using keywords from online medical glossaries specific to each imaging modality or medical domain. To increase the percentage of narrative or educational style videos, a list of priority keywords: "educational", "interpretation", "case study", and similar phrases, are appended to search keywords. We limit our search to channels with <1M subscribers since some channels focus on multiple domains (e.g. radiology channels span CT, MRI, X-ray) and therefore might have a large subscriber base, and channels with >1M subscribers often contain non-imaging videos.

We observe during channel search that searching YouTube for channels by keyword tends to produce irrelevant results, hence, we adopt a video-first search strategy: since video titles are more informative than channel titles, we first find relevant videos, then evaluate the channel of the relevant video for more hits. Each video result is downloaded in low resolution for further analysis. To limit searching irrelevant channels we implement early stopping, wherein, if the first 10 videos of a channel fail the medical filtering step, the channel is skipped, allowing us to keep compute cost low while increasing our pool of visited channels.

\subsection{Medical Filtering}\label{sec:medical_video_filtering}
Each potential pedagogy video is evaluated by the following heuristics:
\begin{enumerate}
\item The video duration is longer than 1 minute and shorter than 2 hours as videos outside this range usually contain little medical imaging information.
\item Video contains speech. We check this either through the video's transcript from the YouTube API, or if not present using the inaSpeechSegmenter \footnote{https://github.com/ina-foss/inaSpeechSegmenter} tool on the first minute of audio. 
\item The number of medical scene frames exceeds the empirically determined threshold unique to each medical domain. This heuristic filters for narrative-style videos (See Section \ref{narrative-filter}).
\end{enumerate}

To expand on the third heuristic, we extract the key-frames of a video for classification; for static domains, we utilize FFmpeg \footnote{https://github.com/FFmpeg/FFmpeg} to detect scenes and extract key-frames (frames with significant visual changes from previous frames). We experiment with scene detection thresholds to determine the optimal threshold per domain across various video durations. For non-static domains, we leverage adaptive content scene detection in \footnote{https://github.com/Breakthrough/PySceneDetect} to avoid capturing frames that are visually different but still part of the same shot (which are characteristic of non-static domains). Camera movements are common in domains such as surgery or endoscopy, and nearly duplicate frames that would be generated by thresholding on video content are instead merged when using PySceneDetect's adaptive detection algorithm. We specifically tune the adaptive detection for each domain by experimentally determining parameters for the algorithm on sample videos from each domain. 

We then classify the key-frames of a video using pre-trained classifiers per domain (see Section \ref{sup:search-classifier-training}). Using the percentage of key-frames predicted to be medical images, videos are differentiated into three categories: positive videos, near-positive videos, and negative videos. For example, a video with 50\% of key-frames predicted to be MRI images is a candidate for further processing, while a video with only 2\% of key-frames is not. Positive videos contain sufficient medical content for the given domain, while near-positive videos may or may not contain sufficient medical content. The thresholds defining positive/near-positive/negative are unique to each domain. We then manually examine a subset of the near-positive category, and determine a more fine-grained percentage threshold to extract more positive videos out of the pool of near-positive videos. See Table \ref{search-thresholds} for the final percentage thresholds used. 

\begin{table}[!ht]
    \centering
    \begin{tabular}{l|c}
    \hline
        \textbf{Domain} & \textbf{Threshold (\%)} \\ \hline
        CT \& X-ray & 10 \\ 
        MRI & 5 \\ 
        Dermatology \& Dentistry & 30 \\ 
        Endoscopy \& Surgery & 50 \\ 
        Ultrasound & 40 \\ 
        Ophthalmology & 35 \\ 
        Mammography & 25 \\ 
        General medical illus. & 20 \\ \hline
    \end{tabular}
    \caption{Final percentage thresholds used during video key-frame classification.}
    \label{search-thresholds}
\end{table}

\subsection{Narrative Filtering}
\label{narrative-filter}
We define narrative-style videos as pedagogy videos where the narrator focuses on describing or analyzing medical images onscreen. To filter for these videos, we first check the first minute of each medical video for speech using inaSpeechSegmenter to ascertain the presence of a narrator. 

For static domains like X-ray, we define a narrative streak as any partition of the video where frames sampled close (w.r.t. time) together are similar using cosine similarity, indicating the narrator is spending a lot of time analyzing that frame. Specifically, we randomly sample a fixed number of clips across each video, sampling three consecutive frames from each clip and checking for similarity. If all three have similarity scores $\geq$ a preset threshold of $0.9$, we count it as a narrative streak. A video is tagged as narrative if a domain-specific preset percentage (\%) of the selected frames exhibit a narrative streak. This simple filtering algorithm helps us reduce the number of videos we process from 748k to 74k videos.

For non-static domains like surgery or ultrasound, consecutive key-frames often exhibit significant change so we instead look for persistent narration around key-frames classified as medical. For example, for ultrasound clips, we extract the times for each consecutive positive key-frame accumulating a sequence of start and end times. Within these time intervals, we determine whether speech exists either through the video's YouTube API transcript or by extracting the audio during the selected time interval and using inaSpeechSegmenter to determine if the segment contains any speech. A video is considered narrative if more than half the key-frames have text for more than a domain-specific number of seconds.




\subsection{Text Extraction using ASR and Text Denoising.}\label{sup:text_denoising}
In line with Quilt-1M \cite{ikezogwo2024quilt} we leverage an open-source ASR model - Whisper \cite{radford2022robust} to transcribe all speech from the selected videos and make sure to account for transcription errors using a similar methodology of finding these types of errors and correcting with a language model. We use the whisper-large-v2 model in the stable-ts library for word-level and sentence-level transcription. As anticipated, this model often misinterprets medical terms, thus requiring the use of post-processing algorithms to minimize its error rates. For this, we adopt a similar methodology proposed in Quilt-1M \cite{ikezogwo2024quilt} to identify, correct, and verify these errors, please see section A.1 in Quilt-1M \cite{ikezogwo2024quilt} supplementary for more details.

\label{med_text_ext}
From the ASR-corrected text, we extract \textit{medical text} which describes the image(s) as a whole. Also, when the speaker describes/gestures at visual regions-of-interest through statements like \textit{"look here ..."}, we extract the text entity being described as \textit{ROI text} in line with Quilt-1M \cite{ikezogwo2024quilt}.

\begin{figure*}[!ht]
    \centering
    \includegraphics[width=\textwidth]{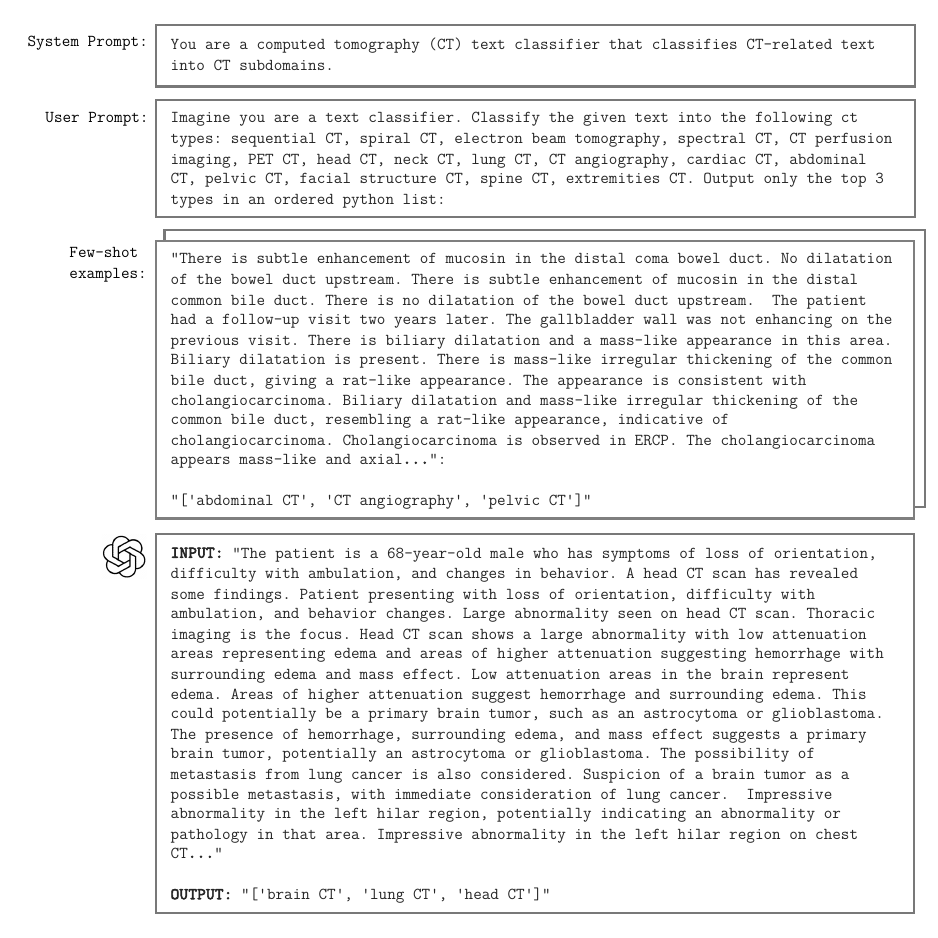} 
    \caption{The GPT-3.5 Turbo prompts used for determining which specific sub-domains are discussed in a video, with few-shot examples.}
    \label{fig:subdomain-prompt}
\end{figure*}

To extract relevant text, we prompt LLMs to filter out all non-medically relevant text, providing context as necessary, while conditioning the LLMs to refrain from introducing new words beyond the corrected noisy text and set the model's temperature to zero. Lastly, the LLM is used to categorize our videos into subdomains by conditioning with a few examples and prompting with the corrected video transcript as input (see Figure \ref{fig:subdomain-prompt} for prompt and sample input/output).

\subsection{Aligning modalities}\label{sec:alignment}
\textbf{Videos}: 
We modify Quilt-LLaVA \cite{seyfioglu2024quilt} pipeline. To align image, text, and trace modalities we compute time chunks for each video denoted as $[(t_1, t_2), (t_3, t_4), \cdots (t_{n-1}, t_n)]$ from key-frames after discriminating for medical frames using the pretrained classifiers -- (\textit{scene\_chunks}). Each \textit{scene\_chunk} is padded with \textit{pad\_time} to its left and right. We use the methods described above to extract the medical/ROI captions as well as the representative image(s) for every chunk/time-interval in \textit{scene\_chunks}
Finally, each chunk in \textit{scene\_chunks} is mapped to text (both medical and ROI captions), traces, and images. Next, we map each image to one or more text (with traces). Using the images' time interval, we extract \textit{raw\_keywords} using the Rake method from the transcript. We extract \textit{keywords} from each medical text returned using the LLM. Finally, if the \textit{raw\_keywords} occur before or slightly after a selected representative image, and overlap with the \textit{keywords} in one of the Medical/ROI texts for that chunk, we map the image to the medical/ROI text. Traces are encoded as the cartesian position of the cursor relative to the image size, we use \( (x^{t}_{j}, y^{t}_{j}) \), where \( x \in [0, W] \) and \( y \in [0, H] \), with \( W \) and \( H \) representing the image width and height, respectively, \( t \) spans from 0 up to the total duration of the \( j^{th} \) stable chunk. 

\noindent \textbf{Articles}: 
The majority of our curated PubMed data uses alphabetic labels in compound figures to denote sub-figures, which increases the complexity of pairing individual sub-figures from compound figures to sub-captions. Our solution leverages an optical character recognition (OCR) model \footnote{https://github.com/JaidedAI/EasyOCR} on each sub-figure to detect the sub-figure labels, which we then match to the extracted sub-caption labels. We impose a 95\% confidence threshold on predicted text to isolate the sub-figure label, as text detected at lower confidence is often non-label text present in the figure (e.g. axis titles, graphs). We then match and pair the detected sub-figure label with the sub-caption label. Despite the generality of this approach, we identified a few failure cases and proposed an error-handling solution in section \ref{sup:sub-sub-pairing} in the Appendix.


\begin{table}[ht!]
  \centering
  \scriptsize
  \begin{tabular}{l|l}
    \toprule
    Hyperparameter              & Training \\
    \midrule
    Batch size (per GPU)        &   256    \\
    Epochs                      &    20   \\
    Peak learning rate          &   1e-5    \\
    Learning rate schedule      &  cosine decay \\
    Warmup (in steps)           &    2000   \\
    Augmentation                &   Resize; RandomCrop (0.8, 1.0)    \\
    Optimizer momentum          &  $\beta_1$, $\beta_2$ = 0.9, 0.98  \\
    Weight decay                &  0.2    \\
    Optimizer                   & AdamW  \\
    
    \bottomrule
  \end{tabular}
  \caption{Training hyperparameters for \model}
  \label{hyperparams}
\end{table}

\begin{table}[ht!]
  \centering
  \scriptsize
  \begin{tabular}{l|l|l}
    \toprule
    Hyperparameter              & ResNet50 & ViT-Small \\
    \midrule
    Batch size (per GPU)        &   256 &  32  \\
    Epochs                      &    10  &  100 \\
    Peak learning rate          &   1e-2 &  1e-3  \\
    Learning rate schedule      &    -   & cosine annealing \\
    \multirow{3}{*}{Augmentation} &  RandomResizedCrop  &  RandomResizedCrop \\
                                  &  (224), &  (384, 0.98, 1.0), \\
                                  &  Resize &  RandomHorizontalFlip \\
    Optimizer momentum          &  0.9 & 0.9  \\
    Weight decay                &  1e-4  & - \\
    Optimizer                   & SGD & SGD \\
    
    \bottomrule
  \end{tabular}
  \caption{Training hyperparameters for domain classifiers.}
  \label{hyperparams-domain-classifiers}
\end{table}

\begin{table*}[!ht]
    \centering
    \scalebox{0.75}{
{\fontsize{8pt}{12pt}\selectfont
    \begin{tabular}{l|p{0.5\linewidth}}
    \hline
        \textbf{Datasets} & \textbf{prompt} \\ \hline
        Pcam \cite{veeling2018rotation}, Nck \cite{kather2018100}, Lc2500 \cite{borkowski2019lung}, Mhist \cite{wei2021petri} &  [
        "a histopathology slide showing \{c\}.", 
        "histopathology image of \{c\}.", 
        "pathology tissue showing \{c\}.",
        "presence of \{c\} tissue on image"
        ] \\ 
        Bach \cite{aresta2019bach}, Skin \cite{kriegsmann2022deep}, Osteo \cite{arunachalam2019viable} & \\
        \hline
        Tcga\_til \cite{tcga-til}, DDI \cite{ddi}, Isic \cite{isic1}, Dental \cite{rahman2024dental} & [
        "\{c\} presented in image", 
        "evidence of \{c\} in image", 
        "an image showing \{c\}"
         ] \\
        Gastrovision \cite{gastrovision}, G1020 \cite{bajwa2020g1020}, Octdl \cite{octdl},  VinDrM \cite{vindrmammonguyen} & \\

        VinDrXR  \cite{nguyen2022vindr}, Dresden \cite{carstens2023dresden} , Radimagenet \cite{mei2022radimagenet}  &  \\
        
    \bottomrule
    \end{tabular}}}
    \caption{Zero-shot classification templates used to evaluate \model's zero-shot capacity across all multiple dataset that constitute the medical benchmark.}
    \label{tab:text_prompts_clf}
\end{table*}

\begin{table*}
    \centering
    \scalebox{0.47}{
{\fontsize{8pt}{12pt}\selectfont
    \begin{tabular}{l|c|c|cc|ccccccc|c|c|cc|cc|c|c|c|cccccccccc|c}  
    \toprule
    \textbf{Models} & \textbf{Models} &   \multicolumn{1}{c|}{\textbf{Xray (mAP)}} &  \multicolumn{2}{c|}{\textbf{CT}} &  \multicolumn{7}{c|}{\textbf{MRI}}  &  \textbf{Mammo (mAP)}  &  \textbf{US}  &  \multicolumn{2}{c|}{\textbf{Optha}}   &   \multicolumn{2}{c|}{\textbf{Derm}}  &  \textbf{Endo} &  \textbf{Surg}  &  \textbf{Dental}  & \multicolumn{10}{c|}{\textbf{Histo}} & {\bf Overall}    \\
                    & vindrXR & lung &  abd  &  af  & brain & hip & knee & abd &  shdr  & spine & vindrM & rad & g1020 & octdl & ddi & isic  & gastro & dresden & dental & til  & pcam & lc\_lung & nck & skin & skin\_tumor & lc\_colon & mhist & bach & osteo &    \\ \midrule
    
    CLIP-ViT-B-32 \cite{radford2021learning} & - & 6.95 & 2.74 & 4.14 & 2.75 & 0.77 & 2.67 & 1.86 & 2.23 & 3.13 & 12.29 & 10.64 & 8.45 & 69.61 & 10.22 & 41.31 & 21.76 & 4.94 & 10.66 & - & 21.33 & 61.81 & 61.55 & 29.20 & 4.47 & 9.84 & 65.57 & 50.67 & 25.25 & 58.85 & 21.63 \\
    
    CLIP-ViT-B-16 \cite{radford2021learning} & - & 6.55 & 1.61 & 3.76 & 3.59 & 6.55 & 3.00 & 1.13 & 1.49 & 4.29 & 14.07 & 10.65 & 5.41 & 31.47 & 6.98 & 60.67 & 13.23 & 1.77 & 11.38 & 2.51 & 20.32 & 51.80 & 47.06 & 21.41 & 5.55 & 13.22 & 79.56 & 52.61 & 23.75 & 43.54 & 18.93 \\
    \hline
    PMC-CLIP \cite{lin2023pmc} & A & 6.80 & 10.89 & 5.73 & 1.19 & 8.31 & 3.24 & 7.90 & 0.11 & 2.00 & 2.55 & 11.86 & 3.72 & 37.06 & 3.78 & 52.44 & 3.77 & 0.86 & 8.15 & 27.85 & 68.92 & 47.06 & 32.66 & 14.37 & 3.86 & 29.66 & 49.96 & 47.39 & 19.50 & 46.20 & 19.23 \\
    \pubmedclip \cite{pubmedclip} & A & 7.40 & 6.06 & 2.04 & 1.08 & 8.95 & 1.38 & 4.07 & 0.54 & 8.50 & 31.21 & 10.45 & 15.94 & 68.33 & 7.56 & 35.52 & 18.19 & 2.05 & 9.72 & 12.19 & 24.55 & 50.38 & 33.33 & 26.45 & 8.07 & 23.01 & 63.66 & 62.74 & 15.25 & 43.63 & 20.77 \\
    \biomed \cite{biomedclip} & A & 10.29 & 6.80 & 2.53 & 3.11 & 12.31 & 2.98 & 4.88 & 0.97 & 6.09 & 11.35 & 10.57 & 58.10 & 29.12 & 18.60 & 51.22 & 20.70 & 3.49 & 16.37 & 16.83 & 37.03 & 71.71 & 71.34 & 49.17 & 24.83 & 40.39 & 84.98 & 38.59 & 44.25 & 46.75 & 27.43 \\
    \hline

    \modelthreetwo & V+A & 10.23 & 14.25 & 1.59 & 2.25 & 23.40 & 3.68 & 9.45 & 1.18 & 2.36 & 24.00 & 10.30 & 44.98 & 66.67 & 20.35 & 62.96 & 19.78 & 2.10 & 15.04 & 26.89 & 23.14 & 70.56 & 81.11 & 48.05 & 28.45 & 49.67 & 93.24 & 55.37 & 37.25 & 55.82 & 31.18 \\
    
    \modelPMB & V+A & 9.90 & 10.36 & 2.30 & 3.95 & 8.80 & 1.80 & 4.33 & 0.98 & 8.38 & 25.04 & 12.06 & 52.98 & 29.22 & 24.56 & 57.01 & 37.63 & 2.08 & 19.30 & 16.83 & 49.63 & 71.90 & 82.05 & 51.05 & 39.32 & 48.34 & 71.68 & 61.82 & 52.00 & 42.44 & 30.96 \\
    
    \model & V+A & 9.66 & 27.35 & 1.38 & 2.75 & 7.52 & 2.61 & 9.93 & 2.80 & 3.10 & 22.59 & 11.12 & 63.48 & 33.53 & 21.22 & 72.26 & 37.10 & 2.38 & 18.97 & 16.44 & 20.34 & 65.90 & 72.37 & 52.16 & 42.37 & 59.87 & 94.16 & 60.59 & 52.50 & 41.43 & \textbf{31.99} \\

    \hline \multicolumn{30}{c}{ \textbf{Data Split Ablation}} \\
    \hline
    \model$^*$ & A & 6.52 & 0.84 & 1.19 & 4.03 & 2.25 & 3.12 & 7.51 & 2.44 & 3.42 & 24.25 & 11.91 & 1.83 & 29.80 & 17.83 & 41.01 & 6.94 & 5.16 & 9.25 & 11.22 & 79.29 & 71.57 & 52.23 & 30.24 & 3.62 & 20.27 & 49.52 & 37.67 & 26.50 & 42.90 & 20.84 \\
    \model$^*$ &  V+A & 8.51 & 27.17 & 1.38 & 1.26 & 19.87 & 1.90 & 10.76 & 2.16 & 3.36 & 8.34 & 10.87 & 55.01 & 34.12 & 30.47 & 73.02 & 42.00 & 1.61 & 13.23 & - & 23.62 & 66.94 & 82.01 & 43.74 & 41.46 & 58.10 & 93.66 & 63.15 & 52.25 & 39.78 & 32.49 \\
    \bottomrule
    \end{tabular}}}  
    
    \caption{\textbf{Expanded Zeroshot Classification Results} shows that our model \model outperforms all other baselines including the out-of-domain CLIP and biomedical vision-language models \biomed and \pubmedclip across the constructed medical benchmark. The benchmark covers all 11 medical domains represented, excluding the non-medical domain of medical illustrations. The metric for X-ray and Mammography is mean average precision while the rest is accuracy.}
    \label{tab:specificzeroshotclf}
\end{table*}

\begin{table}[ht!] 
\centering 
\scriptsize 
\begin{tabular}{lccc} 
\hline 
\textbf{Domain} & \textbf{CT} & \textbf{MRI} & \textbf{X-ray} \\ 
\hline 
Image-text-ROI pairs & 79562 & 82760 & 78983 \\ 
Image-text-ROI-text pairs & 127533 & 112940 & 135242 \\ 
Avg. ROI Text/Image ROI & 3.29 & 2.9 & 3.82 \\ 
Num. ROI Text/Video & 98547.0 & 86798.0 & 85684.0 \\ 
Avg. Words/ROI Text & 10.66 & 9.61 & 12.33 \\ 
Avg. ROI UMLS/Text & 1.47 & 1.48 & 1.47 \\ 
Avg. ROI/Image & 1.6 & 1.38 & 1.74 \\ 
Avg. ROI Text/Chunk & 2.61 & 2.33 & 2.54 \\ 
Unique ROI BBox & 45680 & 35102 & 49157 \\ 
Unique ROI Traces & 11429184 & 4797419 & 10661187 \\ 
Avg. ROI Chunk Duration & 12.85 & 6.19 & 14.42 \\ 
Avg. BBox Height & 319.31 & 204.05 & 312.08 \\ 
Avg. Bbox Width & 538.48 & 281.52 & 506.97 \\ 
\hline 
\end{tabular} 
\caption{Characterization of \dataset \textbf{\textit{image-text-trace}} subset, categorized by individual medical domains. The table provides detailed statistics for each medical modality, including the number of unique images, total dataset duration, ASR error rate, and average image resolution. Note: "ROI" in the table is shorthand for traces.} 
\label{roi-stats} 
\end{table}



\section{\dataset: Article Curation}\label{supp:B}
To curate the PubMed subset of \dataset, we download the PubMed Central Open Access Subset (PMC-OA) \cite{pmc_open_access}, containing 5.47 million articles and filter article figures for the same 12 domains as the YouTube subset of \dataset. Our data curation pipeline for PubMed is as follows:

\begin{enumerate}[label=(\Alph*)]
\item Downloading PMC-OA and extracting each article's XML and images.
\item Parsing each XML to extract figure captions and inline mentions of figures.
\item Filtering for figures with medical imaging with pretrained classifiers.
\item Splitting compounded figures and captions using fine-tuned object detection models and a language model.
\item Pairing correctly split sub-captions and sub-figures together using a combination of optical character recognition (OCR), bounding box heuristics, and error correction.
\item Matching inline mentions of figures with sub-figure/sub-caption pairs using a language model.
\end{enumerate}

In the following sections, we will discuss the \dataset PubMed data pipeline.

\subsection{Caption Extraction}\label{sup:article_caption_extraction}
From each obtained PubMed article, we extract the XML and image files for figure processing. The figure captions are extracted from the paper XML, cleaned, and paired with the corresponding image file of the figure. Additionally, we find all inline mentions of the figure and save them to the figure-caption sample. This yields 23.6M figure-caption samples.

\subsection{Medical Filtering}\label{sup:pubmed_medical_filtering}
To determine whether a figure belongs to one of the twelve domains of \dataset, we train a ResNet-50 CNN for binary classification. We use the same training datasets (see Table \ref{classifier-datasets}) selected when curating \dataset YouTube data, with a binary medical/non-medical label as the target prediction. This filtering step reduces the number of potential figures to 1.03M figures.
To determine the specific domain or domains of each figure, we re-use the medical domain classifiers from the medical filtering step. 

\subsection{Sub-figure Detection}
\label{sup:subfig-detect}

The majority of figures after medical filtering are compound figures, which compress detailed information into a single image and caption. 
Splitting these compound figures into sub-figures is a non-trivial task, since there is no uniform compound figure layout. In contrast to Quilt-1M's \cite{ikezogwo2024quilt} image processing-based approach to splitting these figures, we opt for an object detection approach, which we empirically determined is capable of handling wider range of abstract layouts.

\begin{figure*}[h!]
    \centering
    \begin{subfigure}[b]{0.8\textwidth}
        \centering
        \includegraphics[width=\textwidth]{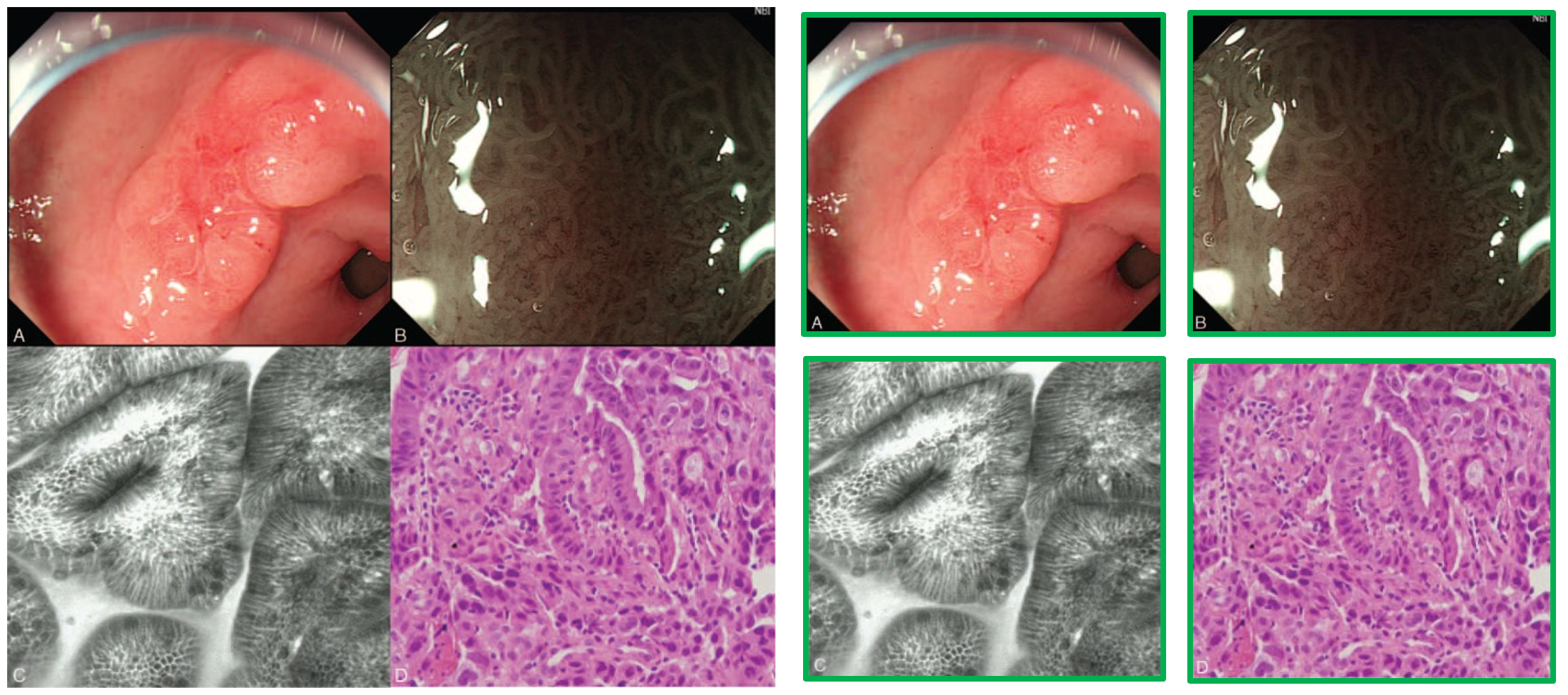}
        \caption{}
        \label{fig:sub_a}
    \end{subfigure}

    \begin{subfigure}[c]{0.37\textwidth}
        \centering
        \includegraphics[width=\textwidth]{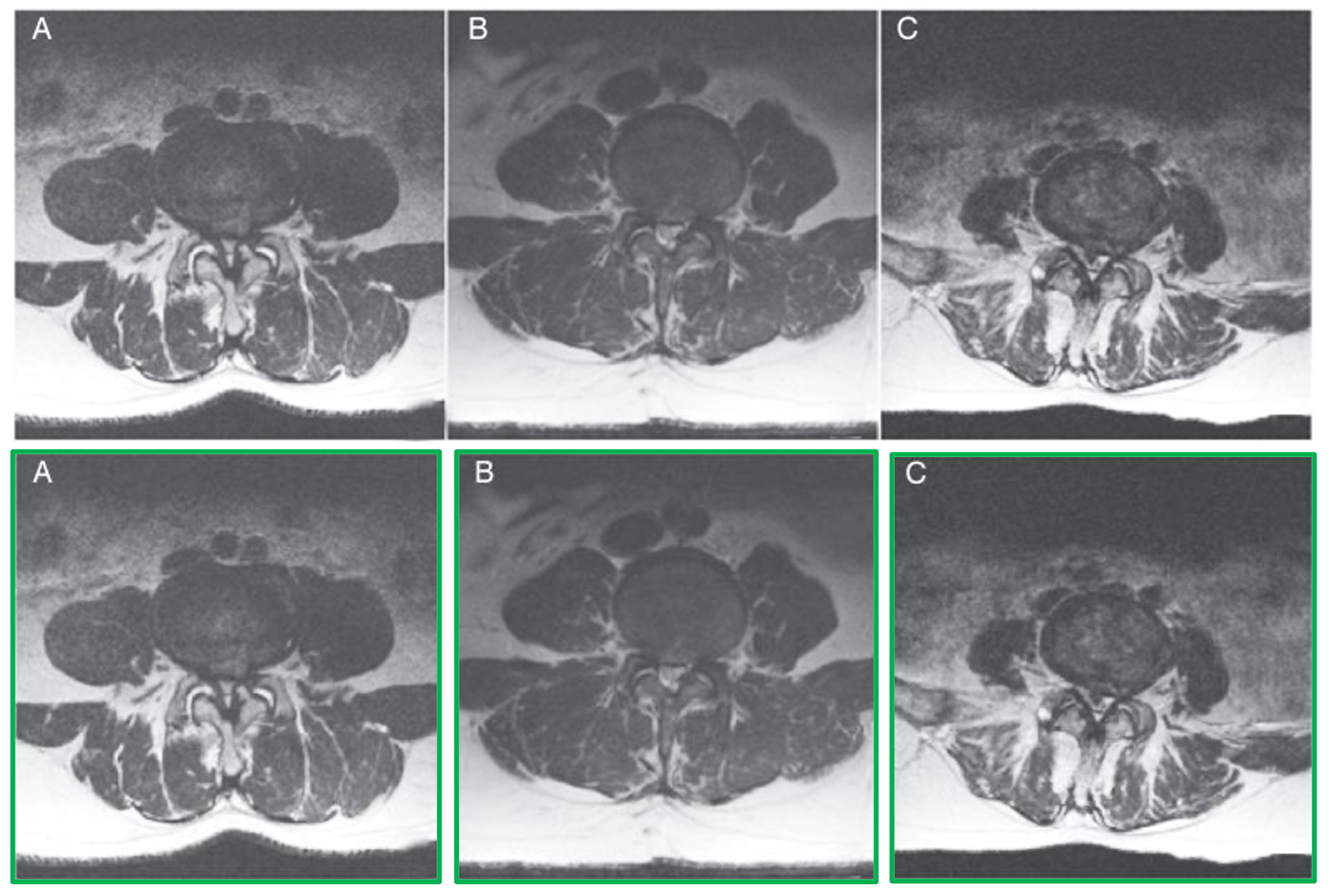}
        \caption{}
        \label{fig:sub_b}
    \end{subfigure}
    \hfill
    \begin{subfigure}[c]{0.58\textwidth}
        \centering
        \includegraphics[width=\textwidth]{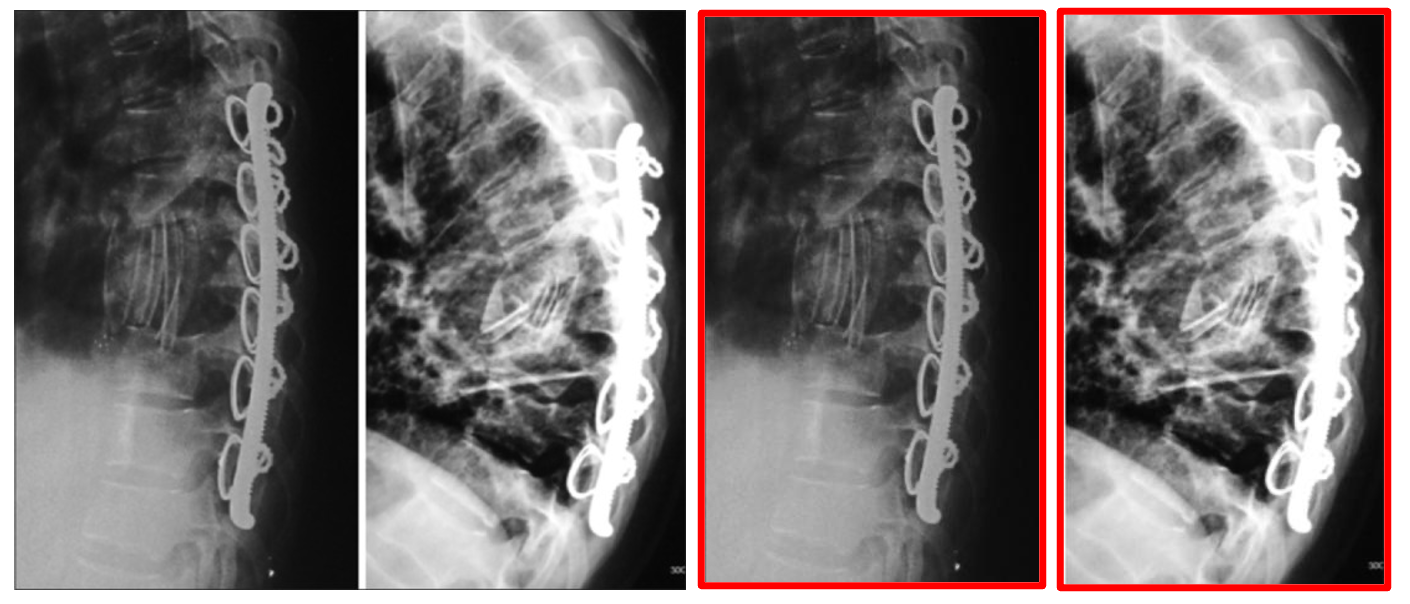}
        \caption{}
        \label{fig:sub_c}
    \end{subfigure}
    
    \caption{(a), (b) Compound figures successfully separated into sub-figures (green), which are then up-scaled and saved. (c) A figure that is incorrectly identified as a compound figure. Since the figure caption contains no sub-captions, the original figure will paired with the entire caption during sub-figure/sub-caption pairing.}
    \label{fig:subfig-detect}
\end{figure*}

Specifically, we finetune a YOLO object detection model \cite{yolov8} to detect sub-figures within compound figures using two medical subfigure separation datasets: MedICaT's sub-figure annotations and ImageCLEF 2016's Figure Separation medical task \cite{medicat, imageclef2016}. MedICaT contains 7507 sub-figure bounding box annotations from 2069 compound figures. ImageCLEF 2016 Figure Separation contains 6782 sub-figure bounding box annotations. We fine-tune a YOLOv8-Large \cite{yolov8} for 100 epochs using an 80/20 training/validation split on the subfigure separation data. One major advantage of the object detection approach is that our model can successfully detect sub-figures even when there is little to no gap/whitespace in-between sub-figures. We process each figure with the fine-tuned YOLOv8 sub-figure detection model, where each detected sub-figure is cropped, and up-scaled by a factor of 4. In the case of compounded figures with uncompounded caption i.e. all constituting images communicate a singular concept (see Figure \ref{fig:subfig-detect}) we pair the caption to the original compounded figure.

\begin{figure*}[h!]
\centering
\includegraphics[width=\textwidth]{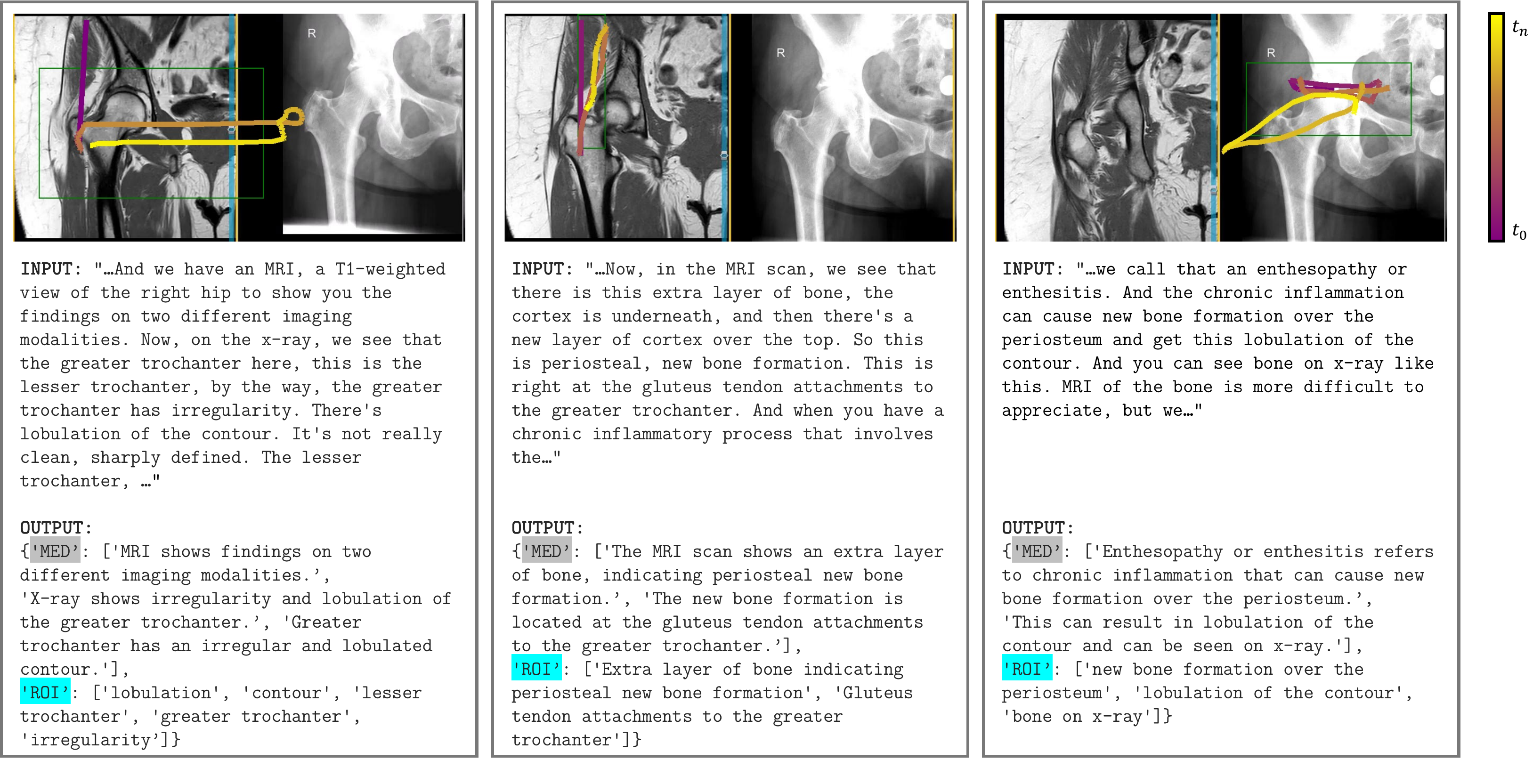}
   \caption{\textbf{\dataset}: Here we show 3 samples from the dataset, these samples come from a single video containing two medical modalities, MRI and X-ray scans, and can be concatenated into an interleaved sample with each sample showing the representative image captured, the raw input text grounded and aligned in-time with the spatial traces \& bbox, and finally the denoised medical and ROI text describing the medical image removing all transcription errors and non-medical information.}
\label{fig:suppmain_data}
\end{figure*}

\subsection{Sub-caption Separation}
\label{sup:subcap-sep}
A compound figure caption usually contains multiple sub-captions. A heuristics-based approach to splitting these compounded captions is difficult to design since figure sub-captions are labeled differently with article authors adhering to varying writing styles typically set by the publishing journal. We therefore opt for an LLM-based approach where we provide diverse examples of sub-caption separation, instructing the language model (GPT-3.5 Turbo) to follow the process below:

\begin{enumerate}
    \item Separate the figure caption into sub-captions based on the sub-figure labels present in the caption e.g. "(A)", "I)", "a.", "bottom left", etc.
    \item Strip the sub-figure labels from each sub-caption text produced.
    \item If any context in the caption pertains to the entire figure, add this context to each sub-caption. This step ensures that each individual sub-caption retains the entire context of the figure.
    \item Return each sub-caption paired with its sub-figure label. 
\end{enumerate}

\begin{figure*}[!ht]
    \centering
    \includegraphics[width=\textwidth]{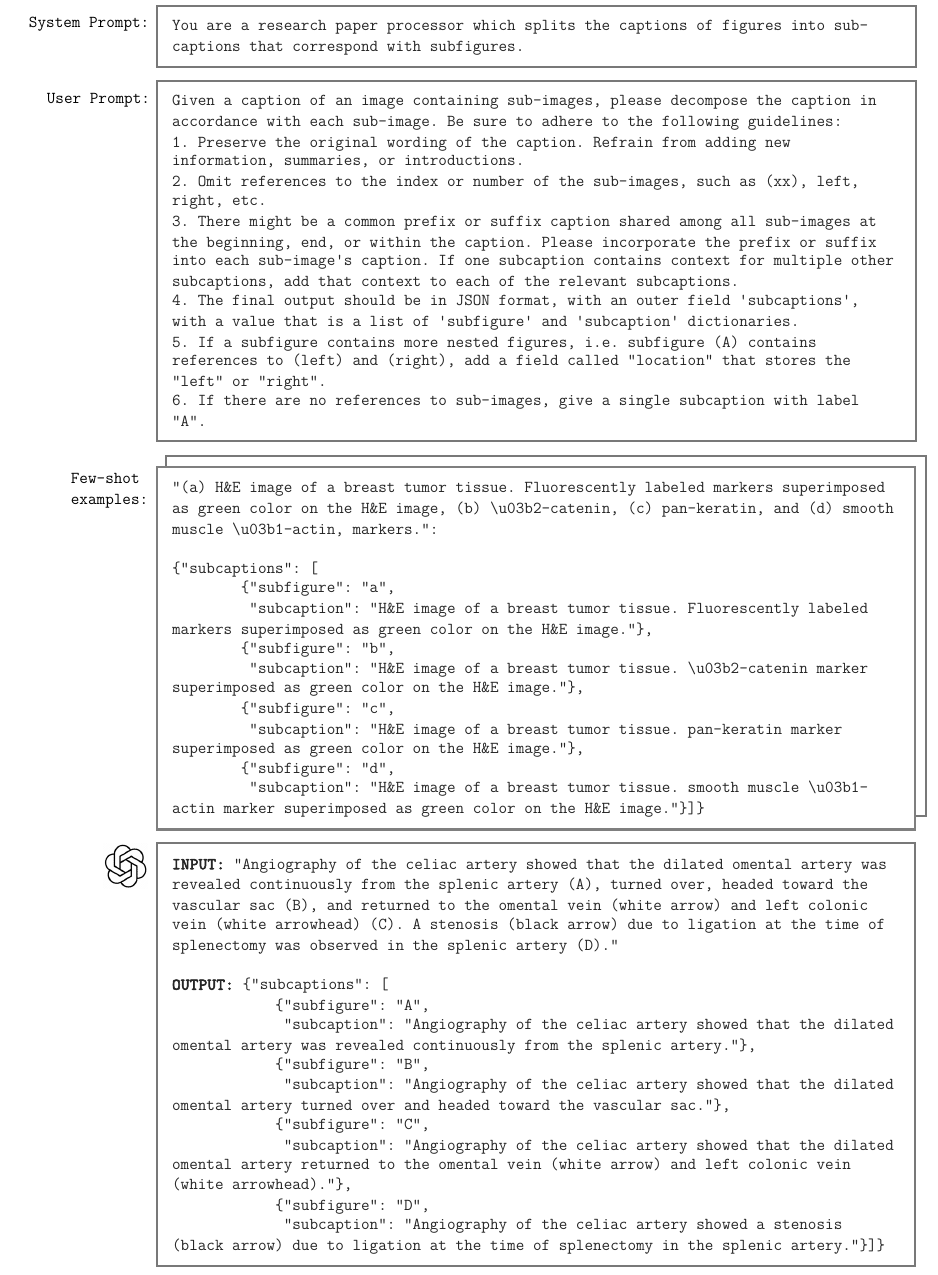} 
    \caption{The GPT-3.5 Turbo prompts used for splitting a compound figure caption into sub-captions, with few-shot examples.}
    \label{fig:caption-split-prompt}
\end{figure*}

We condition the LLM with a few examples, including handling non-compound figures and captions that use spatial cues (e.g. left, center, right) to refer to sub-figures (see full prompt and sample input/output in Figure \ref{fig:caption-split-prompt}). We also process the sub-figure labels returned from the LLM, stripping parentheses and other extraneous characters to make sub-figure/sub-caption pairing easier.

\subsection{Pairing Sub-figures to Sub-captions}
\label{sup:sub-sub-pairing}
Given the separated sub-captions and sub-figures for a compound figure, next we tackle the problem of pairing the correct sub-caption with the correct subfigure. The majority of our curated PubMed data uses alphabetic labels in compound figures to denote sub-figures. Our approach therefore leverages optical character recognition on each sub-figure to detect the sub-figure labels, which we then match to the sub-caption labels extracted during section \ref{sup:subcap-sep}. 

During the sub-figure detection step, we upscale the detected sub-figures by a factor of 4 to enlarge the sub-figure text label for OCR. We impose a 95\% confidence threshold on predicted text during OCR to isolate the sub-figure label. Text detected at lower confidence is often other text in the figure (e.g. axis titles, graphs) being present. We then attempt to match the detected sub-figure label with the sub-caption label. If a match is found, we pair the selected sub-figure and sub-caption. 

There are several types of cases where this approach requires error handling, e.g.:
\begin{enumerate}
\item In a single sub-figure, no labels are identified that exceed the 95\% confidence threshold.
\item Sub-captions use spatial cues to identify sub-figures, e.g. "upper left", "center", "right".
\item If the number of detected sub-figures does not match the number of sub-captions: either some sub-figures or some sub-captions are unpaired.
\end{enumerate}

In case 1, if the compound figure has exactly one sub-figure and one sub-caption left unpaired, we pair the two. Otherwise, we lower the confidence threshold to 80\% and re-detect sub-figure labels, then re-match with sub-captions. Sub-figures that fall in this category tend to have their label close to the border of the cropped sub-figure, have small sub-figure text, or have backgrounds that resemble the font color of the label. For case 2, we use the bounding box coordinates of the detected sub-figures and the spatial cues provided in the caption to pair figures and captions. For example, a sub-caption with the label "upper left" will be paired with the sub-figure with the upper leftmost bounding box. Lastly, case 3 occurs when either sub-figure detection and/or sub-caption separation perform incorrectly. The majority of figures in this category occur when sub-figure detection identifies multiple sub-figures, but the figure caption contains no sub-captions. In this case, we pair the original figure and caption. 
\begin{table*}[ht!]
    \centering
    \scriptsize
    \begin{tabular}{llllllllllll}
    \hline
        \textbf{Domain} & \textbf{CT} & \textbf{MRI} & \textbf{Endo} & \textbf{Genmed} & \textbf{Surgery} & \textbf{Optha} & \textbf{Mammo} & \textbf{Derma} & \textbf{Ultrasound} & \textbf{X-ray} & \textbf{Dental} \\ \hline
        Unique images & 47441 & 55784 & 43230 & 23985 & 75312 & 758 & 288 & 19639 & 69835 & 54732 & 15375 \\
        Image-text pairs & 89036 & 97065 & 135108 & 54684 & 186807 & 681 & 42 & 27182 & 140251 & 101215 & 27391 \\
        Total Med UMLS & 295064 & 262277 & 504753 & 224338 & 656811 & 3243 & 251 & 106291 & 542695 & 270191 & 57086 \\
        Avg. Med Text/Image & 2.44 & 2.14 & 3.49 & 2.79 & 2.85 & 2.48 & 1.40 & 2.02 & 2.61 & 2.56 & 2.54 \\
        Num. Med Text/Video & 80356 & 73048 & 134385 & 54635 & 185223 & 681 & 59 & 26904 & 139792 & 71778 & 14867 \\
        Avg. Words/Med Text & 28.35 & 24.01 & 40.30 & 36.01 & 32.10 & 36.74 & 15.62 & 23.49 & 31.29 & 30.04 & 30.61 \\
        Avg. Med UMLS/Text & 3.74 & 3.68 & 3.83 & 4.27 & 3.57 & 6.07 & 4.25 & 3.89 & 4.04 & 3.83 & 3.87 \\
        Total Chunks & 41870 & 40133 & 42770 & 23947 & 74188 & 758 & 283 & 19432 & 69486 & 37048 & 7661 \\
        Avg. Chunk Duration & 30.88 & 18.26 & 47.03 & 36.35 & 28.85 & 74.60 & 2.42 & 23.71 & 31.14 & 34.05 & 50.42 \\
        Avg. Med Text/Chunk & 2.12 & 1.97 & 3.23 & 2.35 & 2.48 & 1.27 & 0.16 & 1.00 & 1.80 & 2.07 & 1.69 \\
        Avg. Images/Chunk & 1.27 & 1.51 & 1.01 & 1.00 & 1.03 & 1.00 & 1.02 & 1.01 & 1.01 & 1.75 & 2.03 \\
        Avg. Image-Text/Chunk & 2.55 & 2.66 & 3.25 & 2.35 & 2.52 & 1.27 & 0.12 & 1.01 & 1.81 & 3.36 & 3.18 \\
        Precision (Unconditioned) & 0.16 & 0.15 & 0.18 & 0.17 & 0.18 & 0.20 & 0.33 & 0.23 & 0.16 & 0.19 & 0.20 \\
        Precision (Conditioned) & 0.49 & 0.45 & 0.48 & 0.56 & 0.54 & 0.44 & 0.73 & 0.43 & 0.40 & 0.46 & 0.42 \\
        Clinical ASR Error Rate & 0.01 & 0.01 & 0.01 & 0.01 & 0.01 & 0.01 & 0.01 & 0.01 & 0.01 & 0.01 & 0.01 \\
        Total Duration (hrs) & 327.0 & 416.0 & 428.0 & 281.0 & 389.0 & 15.0 & 0.0 & 187.0 & 1182.0 & 562.0 & 140.0 \\
        Avg. Duration (mins) & 13.27 & 12.07 & 6.24 & 9.75 & 7.63 & 6.94 & 0.11 & 8.05 & 8.58 & 19.24 & 13.68 \\
        Total ASR len. (words) & 2355609 & 3364120 & 2722335 & 2085480 & 2975575 & 89011 & 4399 & 1204875 & 9481084 & 4476475 & 1149722 \\
        Avg. ASR len. (words) & 1592.70 & 1624.39 & 660.76 & 1204.09 & 972.73 & 659.34 & 879.80 & 863.09 & 1146.72 & 2550.70 & 1866.43 \\
        \hline
    \end{tabular}
    \caption{Characterization of \dataset \textbf{\textit{image-text}} subset, categorized by individual medical domains. The figure provides detailed statistics for each medical modality, including the number of unique images, total dataset duration, ASR error rate, and average image resolution.}
    \label{video-stats}
\end{table*}


\subsection{Inline Figure Reference Pairing}
\label{sup:inlinepairing}
\begin{figure*}[ht!]
    \centering
    \includegraphics[width=\textwidth]{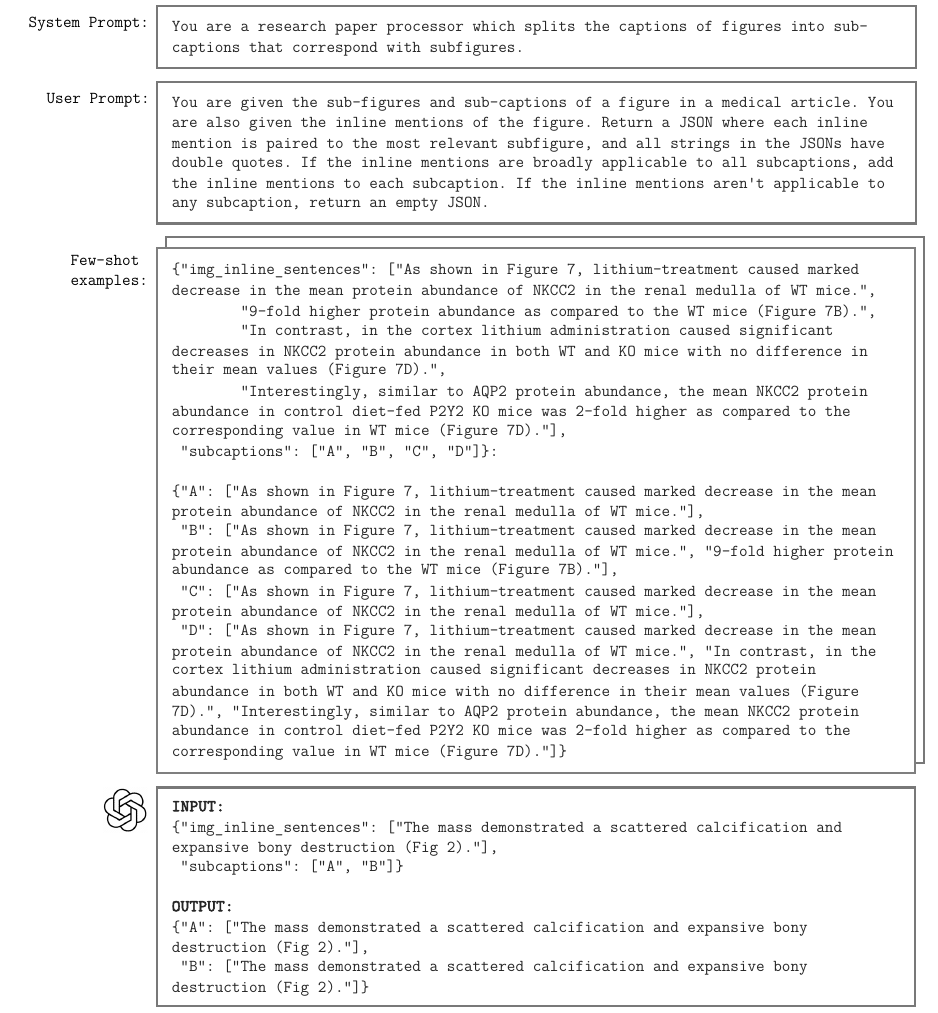} 
    \caption{The GPT-3.5 Turbo prompts used for pairing inline references of a figure with the most relevant sub-figures, with few-shot examples.}
    \label{fig:inline-pair-prompt}
\end{figure*}

In the final step of the pipeline, we pair the inline reference of a figure with the figure caption since inline references contain valuable context about the figure. However, an inline reference may refer to a sub-figure instead of the entire figure. We therefore utilize a language model to determine which sub-figure is most relevant to an inline reference. For each sample, we prompt an LLM (GPT-3.5 Turbo) with the list of sub-figure labels and a list of inline references and task the model with determining which sub-figure label best corresponds to the inline reference. In the case that the reference cites the entire figure instead of a sub-figure, we consider the inline reference relevant to all sub-figures. For each relevant sub-figure, we add the inline reference to its list of captions. See Figure \ref{fig:inline-pair-prompt} for the complete prompt and sample input/output. 

\section{Characterizing \dataset}
\label{xrizing_dataset_full}

To create \dataset we combine medical narratives curated from videos with image-text pairs curated from PubMed, resulting in 4.7M total image-text samples of which 1M samples are localized narratives. Section \ref{xrizing_dataset} gives an overview characterization of the entire dataset, and Tables \ref{video-stats}, \ref{roi-stats} below provide additional specific characterization details split per domain. Note we omit characterization for Histopathology in the tables below as the details for the domain can be found in prior work. 


\section{Training, Benchmark, and Evaluation}
\label{sup:training_details}

\subsection{\model Training}

We leverage OpenCLIP \cite{ilharco_gabriel} to train our models as it allows us to quickly import our datasets and adapt varying components of our model including the underlying image and text towers and the training hyperparameters. Our experiments utilize Pytorch on 4 NVIDIA L40s GPUs, as well as gradient checkpointing, automatic mixed precision with bfloat16 to reduce memory usage. All other hyperparameters used are listed in Table \ref{hyperparams}. Our dataset is split into 16 tar files in the WebDataset \footnote{https://github.com/webdataset/webdataset} format for training.

\subsection{Benchmarking on Downstream Medical Tasks}
 We evaluate the utility of \model on a new medical imaging benchmark of all medical domains represented in our pre-training dataset \dataset, with some domains represented by $>=1$ dataset/task for classification, totaling 29 downstream datasets and on a held-out set of 1000 unique images for the retrieval task downstream.
For MRI we use the \textbf{RadImageNet} \cite{mei2022radimagenet} MRI subsets tasks based on the anatomical region scanned in the image these include Ankle/foot with 25 classes, Brain with 10 classes, Knee with 18 classes, Abdomen/pelvis with 26 classes, Hip with 14 classes, Shoulder with 14 classes, Spine with 9 classes. To evaluate on CT domain we also use RadImageNet's \cite{mei2022radimagenet} CT dataset which cover two (2) anatomical regions with Lung having 6 sub-classes and Abdomen/pelvis with 28 subclasses. For ultrasound, we evaluate on RadImageNet's \cite{mei2022radimagenet} US dataset which covers a total of 15 classes across Thyroid and Abdomen/pelvis anatomical regions. For Xray, we evaluate on \textbf{VinDr-CXR} Chest Xrays \cite{nguyen2022vindr} test set and report the mean average precision (mAP) across all 28 findings, similarly to evaluate on Mammography we use \textbf{VinDr-Mammo} \cite{vindrmammonguyen} and report the mAP on all X findings, leveraging only the standard bilateral craniocaudal (CC) view of the test set. We evaluate on surgical organ classification using \textbf{Dresden} \cite{carstens2023dresden} which covers 8 abdominal organs; to evaluate for endoscopy domain we test on all procedures images in \textbf{GastroVison} \cite{gastrovision} with 27 classes. For Dermatology we evaluate on the \textbf{Diverse Dermatology Images} (DDI) \cite{ddi} binary (benign or malignant) dataset and Isic 2018 dataset \cite{isic1}.
For Dentistry we evaluate on \textbf{Dental orthopantomography} (OPG) \cite{rahman2024dental} X-ray dataset with 6 classes. To evaluate the Ophthalmology domain we evaluate on \textbf{G1020} \cite{bajwa2020g1020} a retinal fundus glaucoma dataset and on \textbf{Optical Coherence Tomography Dataset} (OCTDL) \cite{octdl} with 6 disease classes. We evaluate the Histopathology domain on the following datasets: \textbf{PatchCamelyon} \cite{veeling2018rotation} for lymph node metastatic tissue binary prediction task, \textbf{NCT-CRC-HE-100K} \cite{kather2018100} on 8 morphological classes, \textbf{BACH} \cite{aresta2019bach} which consists of breast tissues with 4 classes including being and invasive carcinoma, \textbf{Osteo} \cite{arunachalam2019viable} osteosarcoma dataset with 3 classes including necrotic tumor, \textbf{SkinCancer} \cite{kriegsmann2022deep} dataset of tissue patches from skin biopsies of 12 anatomical classes and 4 neoplasm categories that make up the SkinTumor Subset, we also evaluate on \textbf{MHIST} \cite{wei2021petri} dataset of colorectal polyps tissue, \textbf{LC25000} \cite{borkowski2019lung} dataset, which is split in-between LC25000 (Lung) and LC25000 (Colon), for lung and colon adenocarcinomas classification, and on TCGA-TIL \cite{tcga-til} for tumor-infiltrating lymphocytes (TILs) binary classification, based on H\&E images from 13 of The Cancer Genome Atlas (TCGA) tumor types.

\subsection{Evaluation}
To evaluate zero-shot classification capacity across all constituting datasets in our medical benchmark outlined in \ref{sec:training} we leverage simple prompts listed in Table \ref{tab:text_prompts_clf}, with the specific results shown in Table \ref{tab:specificzeroshotclf}.

\subsection{Search Classifiers}\label{sup:search-classifier-training}

To classify images into domains, we train a ResNet50 for 10 epochs and a ViT-Small for 100 epochs using DINO on a binary classification task for each medical domain. Both types of models are trained on 4 NVIDIA A4000 GPUs. All hyperparameters are listed in Table \ref{hyperparams-domain-classifiers}. For each classifier, we use domain-specific datasets as positive samples and non-medical datasets as negative samples. For the binary medical/non-medical classifier used in Section \ref{sup:pubmed_medical_filtering}, we use all medical domain datasets as positive samples, and the same group of non-medical datasets as negative samples. See Table \ref{classifier-datasets} for an overview of the datasets used to train these classifiers.

\section{\dataset Examples}
\label{supp:examples}
Below, we show examples in the dataset across all 12 modalities and representative examples of the types of interleaved samples within the dataset.

\begin{figure*}[ht!]
    \centering
    \includegraphics[width=\textwidth]{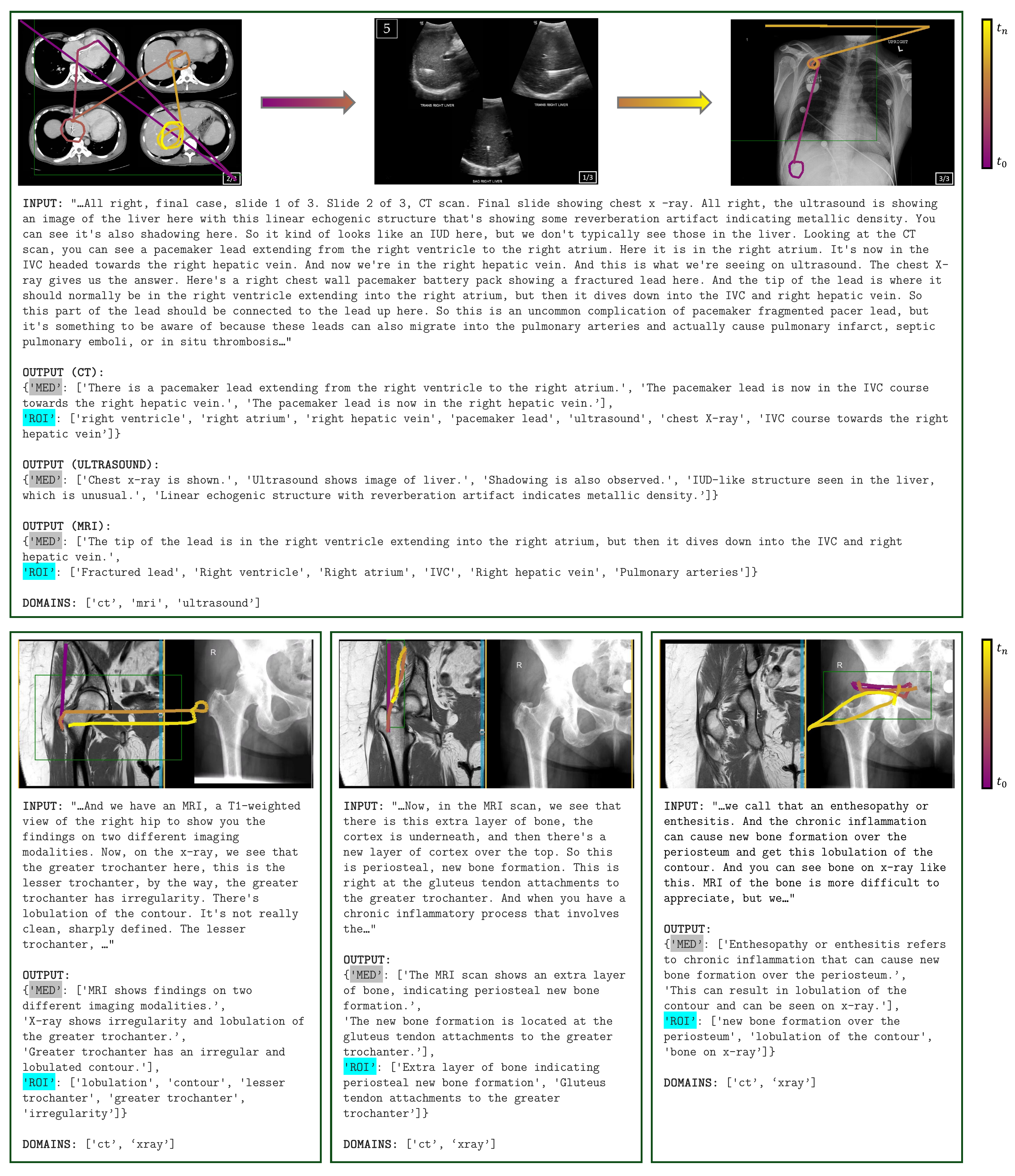} 
    \caption{\textbf{ Interleaved examples} within the \dataset dataset. \textbf{Input}: raw input text from ASR. \textbf{Output}: the output from the LLM, with denoised medical text corresponding to different modalities. \textbf{Domains}: classification of the sample into domains.}
    \label{fig:interleaved-examples}
\end{figure*}

\begin{figure*}[ht!]
    \centering
    \includegraphics[width=\textwidth]{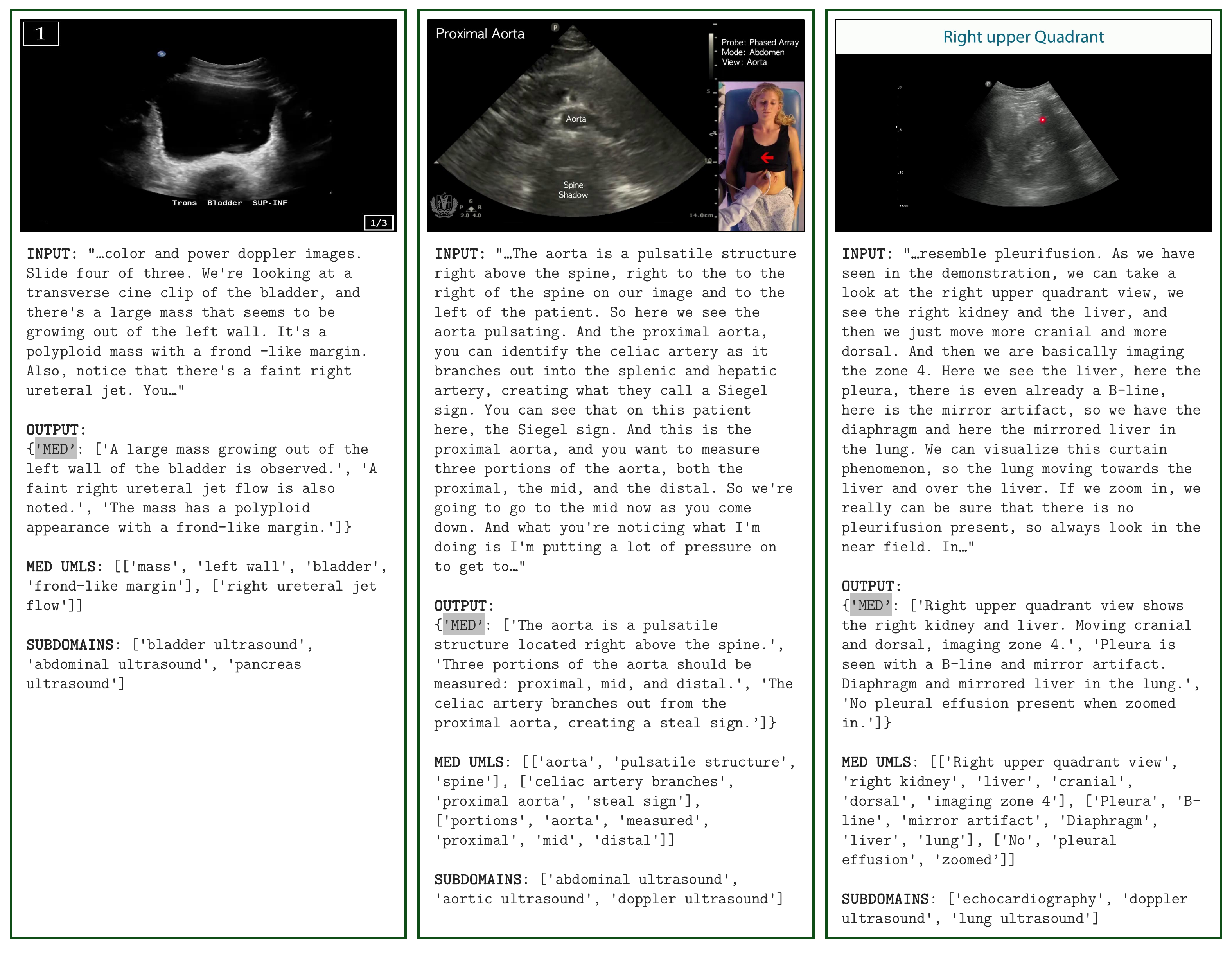} 
    \caption{\textbf{ Ultrasound examples} within the \dataset dataset. \textbf{Input}: raw input text from ASR. \textbf{Output}: the output from the LLM, with denoised medical. \textbf{UMLS}: UMLS entities extracted from the medical text. \textbf{Subdomain}: classification of the sample into finer-grained subdomains.}
    \label{fig:ultra-examples}
\end{figure*}

\begin{figure*}[ht!]
    \centering
    \includegraphics[width=\textwidth]{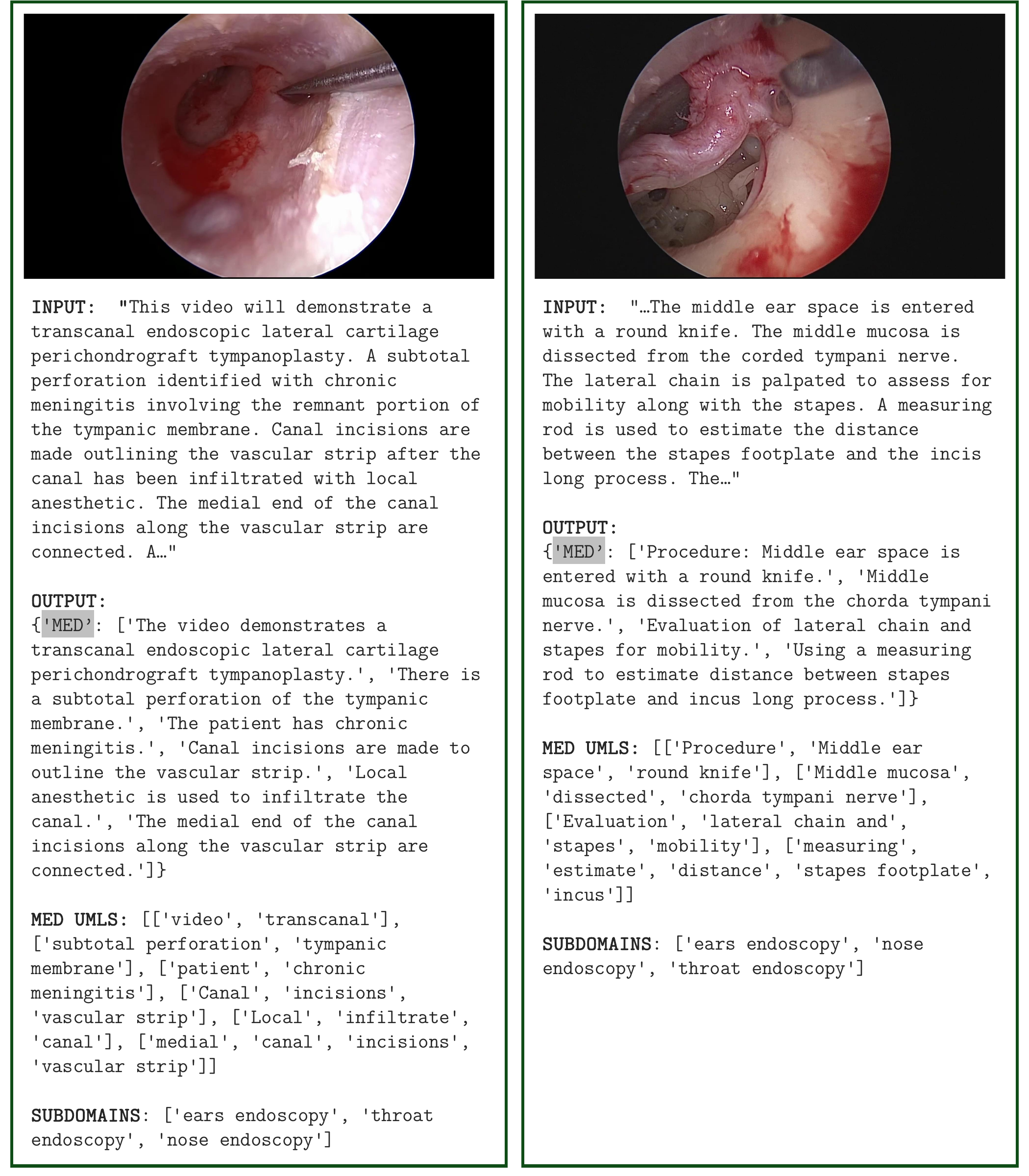} 
    \caption{\textbf{ Endoscopy examples} within the \dataset dataset. \textbf{Input}: raw input text from ASR. \textbf{Output}: the output from the LLM, with denoised medical. \textbf{UMLS}: UMLS entities extracted from the medical text. \textbf{Subdomain}: classification of the sample into finer-grained subdomains.}
    \label{fig:endo-examples}
\end{figure*}

\begin{figure*}[ht!]
    \centering
    \includegraphics[width=\textwidth]{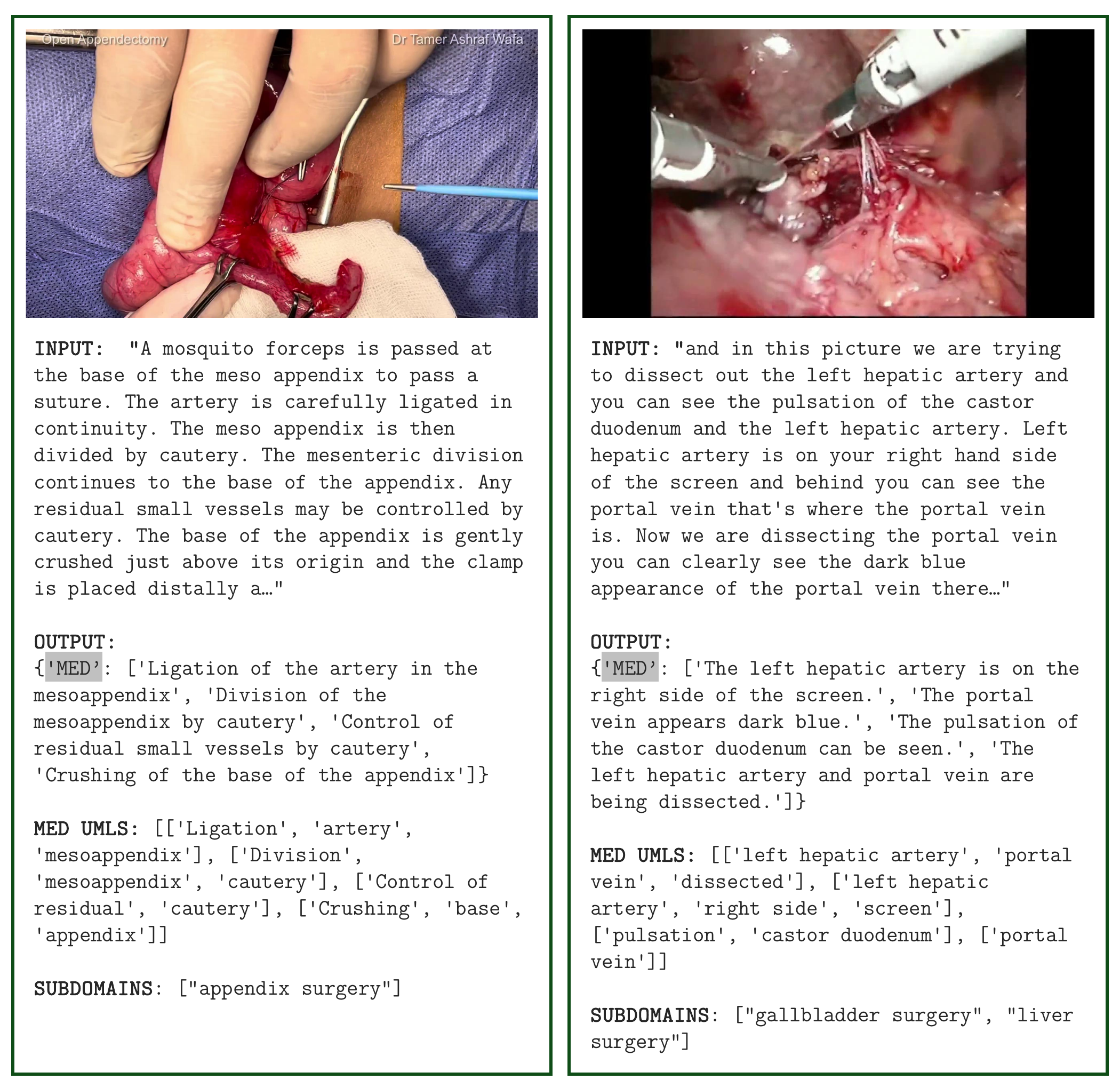} 
    \caption{\textbf{ Surgery examples} within the \dataset dataset. \textbf{Input}: raw input text from ASR. \textbf{Output}: the output from the LLM, with denoised medical. \textbf{UMLS}: UMLS entities extracted from the medical text. \textbf{Subdomain}: classification of the sample into finer-grained subdomains.}
    \label{fig:surgery-examples}
\end{figure*}

\begin{figure*}[ht!]
    \centering
    \includegraphics[width=\textwidth]{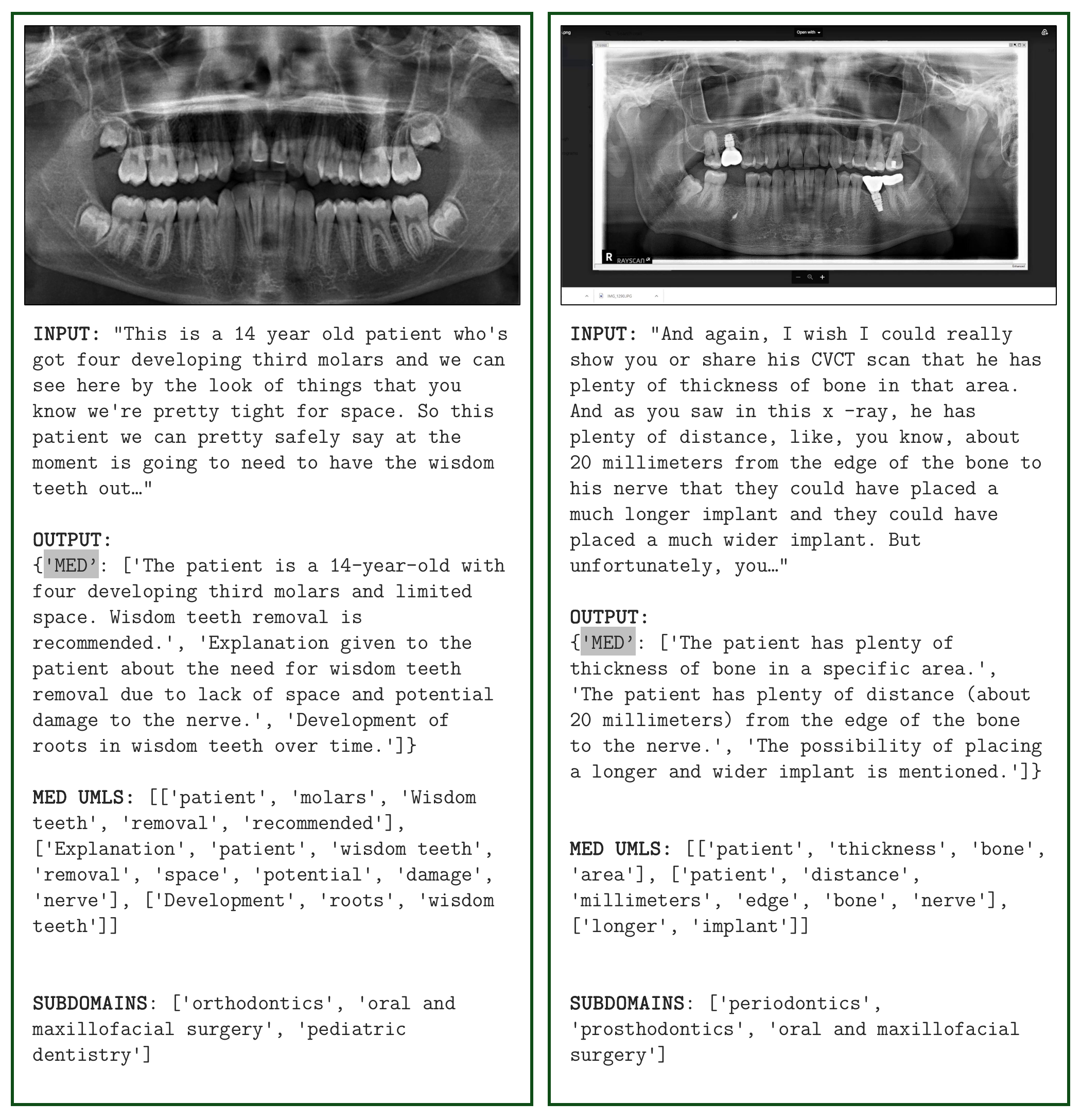} 
    \caption{\textbf{ Dentistry examples} within the \dataset dataset. \textbf{Input}: raw input text from ASR. \textbf{Output}: the output from the LLM, with denoised medical. \textbf{UMLS}: UMLS entities extracted from the medical text. \textbf{Subdomain}: classification of the sample into finer-grained subdomains.}
    \label{fig:dentist-examples}
\end{figure*}

\begin{figure*}[ht!]
    \centering
    \includegraphics[width=\textwidth]{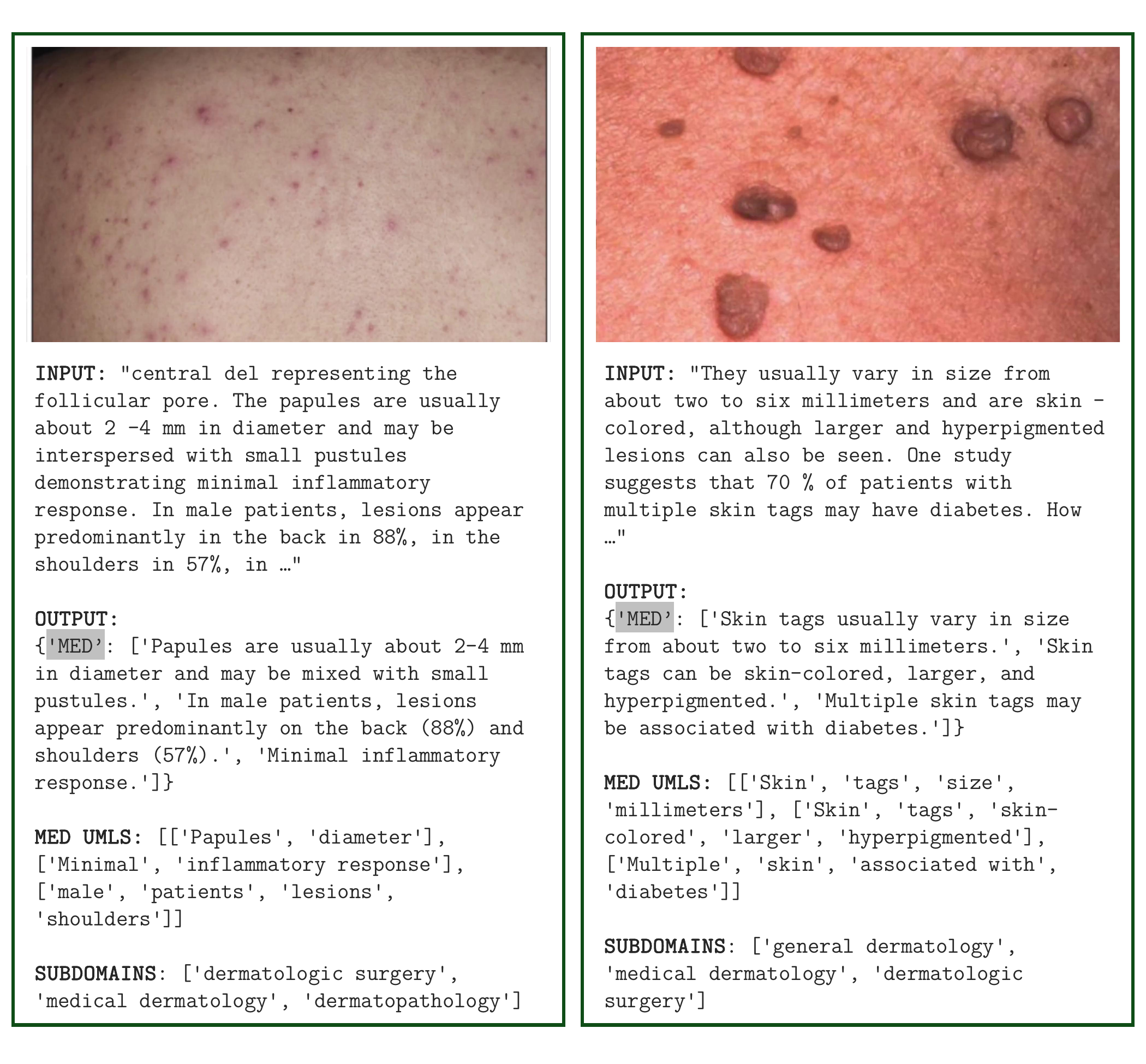} 
    \caption{\textbf{ Dermatology examples} within the \dataset dataset. \textbf{Input}: raw input text from ASR. \textbf{Output}: the output from the LLM, with denoised medical. \textbf{UMLS}: UMLS entities extracted from the medical text. \textbf{Subdomain}: classification of the sample into finer-grained subdomains.}
    \label{fig:derma-examples}
\end{figure*}

\begin{figure*}[ht!]
    \centering
    \includegraphics[width=\textwidth]{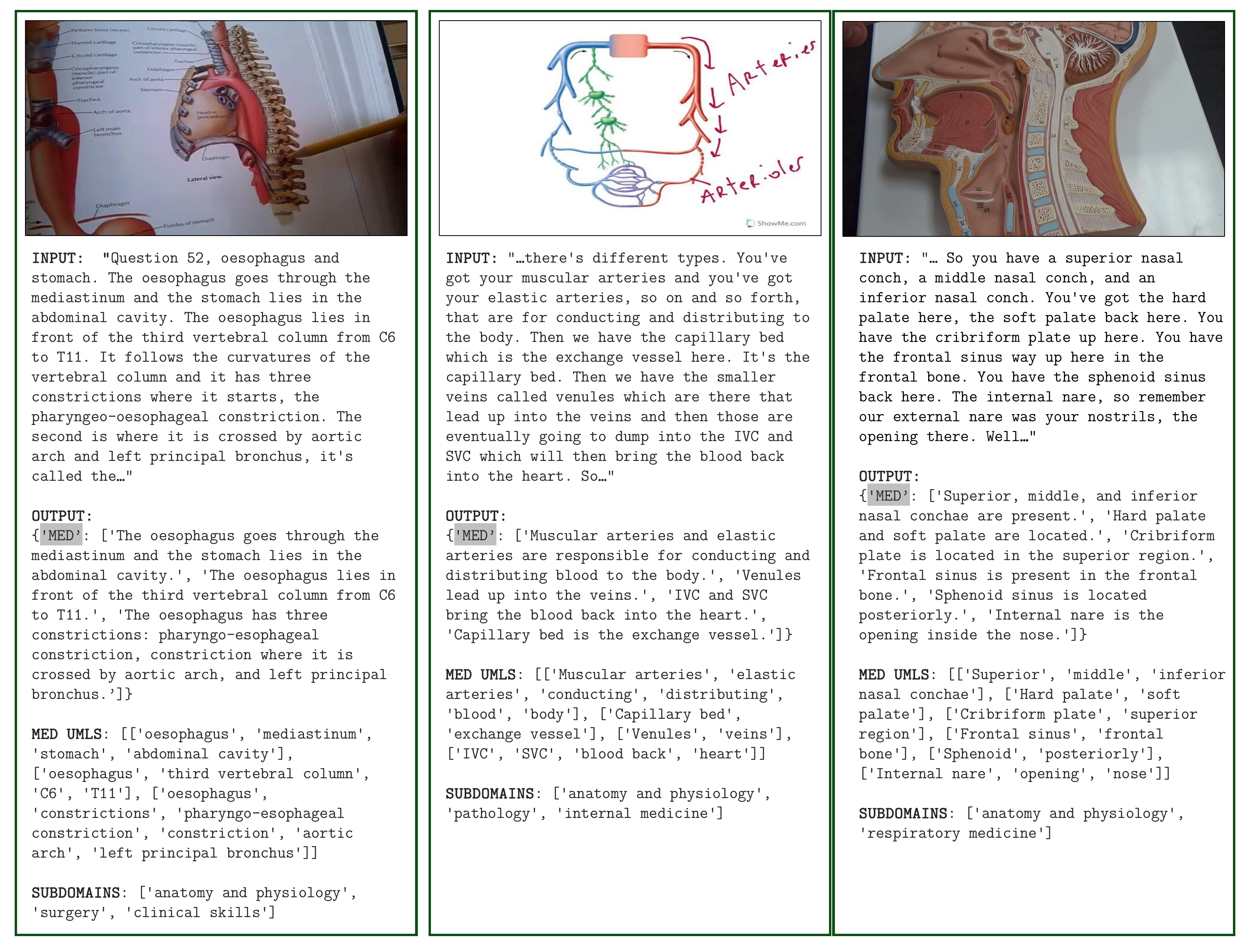} 
    \caption{\textbf{ General medical examples} within the \dataset dataset. \textbf{Input}: raw input text from ASR. \textbf{Output}: the output from the LLM, with denoised medical. \textbf{UMLS}: UMLS entities extracted from the medical text. \textbf{Subdomain}: classification of the sample into finer-grained subdomains.}
    \label{fig:genmed-examples}
\end{figure*}

\begin{figure*}[ht!]
    \centering
    \includegraphics[height=0.7\textwidth]{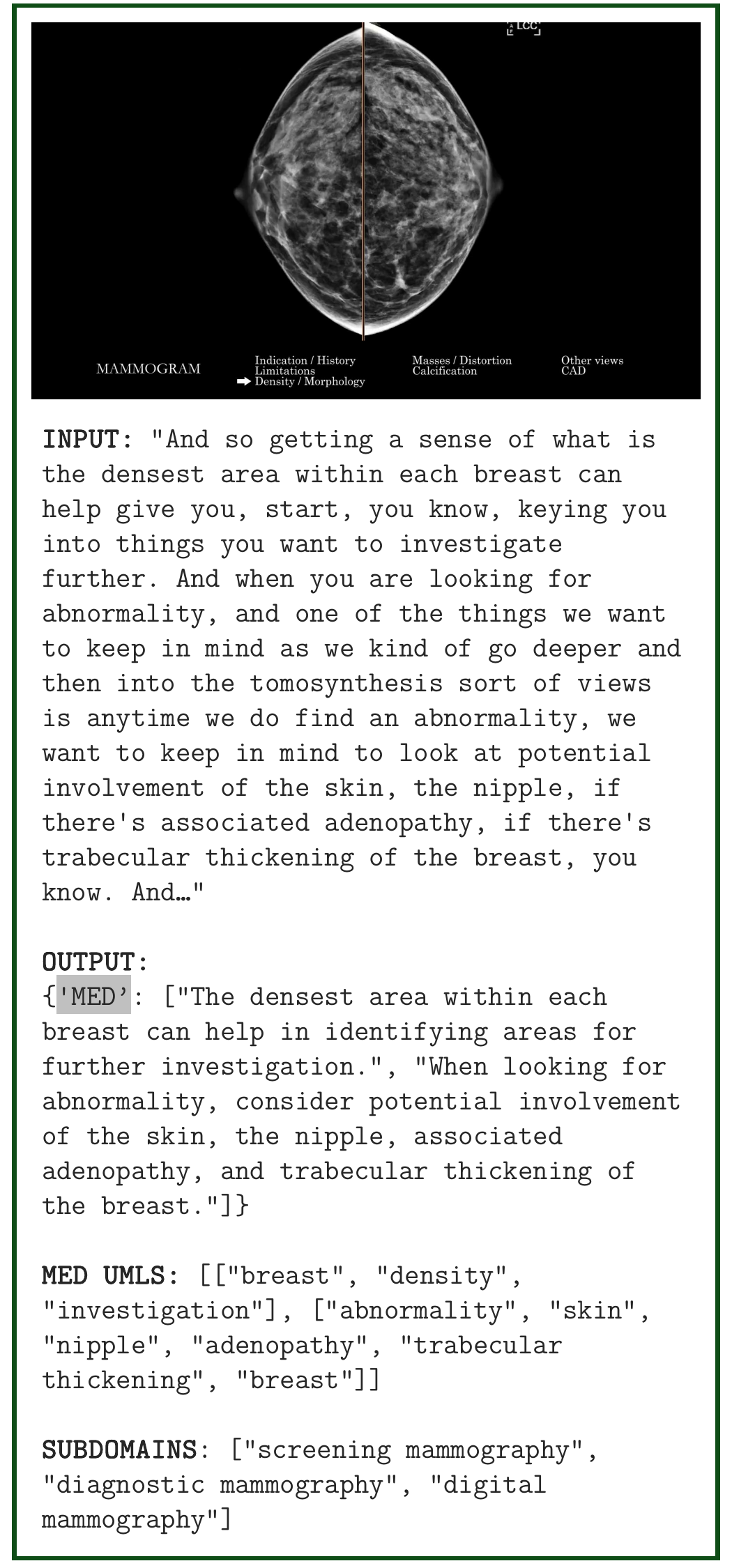} 
    \caption{\textbf{ Mammography examples} within the \dataset dataset. \textbf{Input}: raw input text from ASR. \textbf{Output}: the output from the LLM, with denoised medical. \textbf{UMLS}: UMLS entities extracted from the medical text. \textbf{Subdomain}: classification of the sample into finer-grained subdomains.}
    \label{fig:mammo-examples}
\end{figure*}

\begin{figure*}[ht!]
    \centering
    \includegraphics[height=0.7\textwidth]{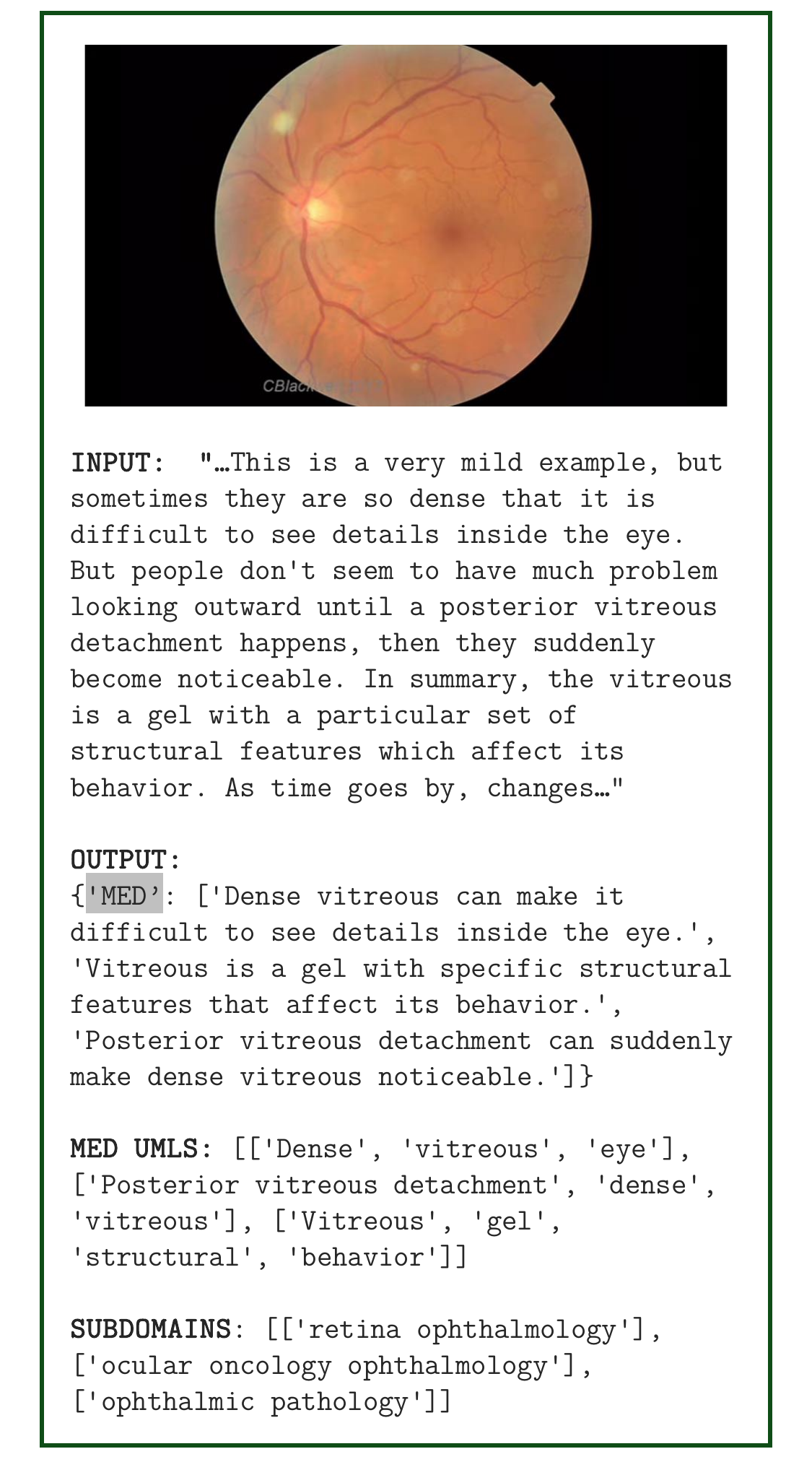} 
    \caption{\textbf{ Ophthalmology examples} within the \dataset dataset. \textbf{Input}: raw input text from ASR. \textbf{Output}: the output from the LLM, with denoised medical. \textbf{UMLS}: UMLS entities extracted from the medical text. \textbf{Subdomain}: classification of the sample into finer-grained subdomains.}
    \label{fig:optha-examples}
\end{figure*}

\begin{table*}[!hb]
    \centering
    \scriptsize
    \begin{tabular}{lc|cccc}
    \hline
        \textbf{Domain} & \textbf{Dataset} & \textbf{Total samples} & \textbf{Train} & \textbf{Test} & \textbf{Image Size} \\ \hline
        CT & LIDC-IDRI \cite{lidc} & 10005 & 7004 & 3002 & 512 $\times$ 512 \\ 
        ~ & TCGA-LUAD \cite{tcga-luad} & 48931 & 34252 & 14679 & 512 $\times$ 512 \\ 
        ~ & WORD \cite{word} & 30495 & 21347 & 9149 & 512 $\times$ 512 \\ 
        ~ & Positive Videos & 1612 & 1128 & 484 & ~ \\ 
        X-ray & ChestX-ray14 \cite{chestxray14} & 112120 & 78484 & 33636 & 1024 $\times$ 1024 \\ 
        ~ & GRAZPEDWRI-DX \cite{grazpedwridx} & 20327 & 14229 & 6098 & 660 $\times$ 1660 \\ 
        ~ & Shoulder X-ray Classification \cite{shoulderxray} & 841 & 589 & 252 & ~ \\ 
        ~ & Digital Knee X-ray \cite{kneexray} & 1650 & 1155 & 495 & 300 $\times$ 162 \\ 
        ~ & MURA \cite{mura} & 40561 & 28393 & 12168 & 1500 $\times$ 2000 \\ 
        ~ & Positive Videos & 692 & 484 & 208 & 1440 $\times$ 1080 \\ 
        MRI & fastMRI \cite{fastmri} & 58847 & 41193 & 17654 & 320 $\times$ 320 \\ 
        ~ & Duke-Breast-Cancer-MRI \cite{dukemri} & 922 & 645 & 277 & 256 $\times$ 256 \\ 
        ~ & Medical Segmentation Decathlon \cite{medsegdecathlon} & 2633 & 1843 & 790 & 256 $\times$ 256 \\ 
        ~ & Positive Videos & 118 & 83 & 35 & ~ \\ 
        Dermatology & Dermnet \cite{dermnet} & 19500 & 13650 & 5850 & 720 $\times$ 472 \\ 
        ~ & DDI \cite{ddi} & 656 & 459 & 197 & 300 $\times$ 300 \\ 
        ~ & 7-point \cite{7point} & 2045 & 1432 & 614 & 480 $\times$ 720 \\ 
        ~ & ISIC \cite{isic} & 33126 & 23188 & 9938 & ~ \\ 
        ~ & Fitzpatrick 17k \cite{fitzpatrick17k} & 16577 & 11604 & 4973 & ~ \\ 
        ~ & HAM10000 \cite{ham10000} & 10015 & 7011 & 3005 & 800 $\times$ 600 \\ 
        Endoscopy & KVASIR \cite{kvasir} & 8000 & 5600 & 2400 & 720 $\times$ 576 \\ 
        ~ & ITEC LapGyn4 \cite{iteclapgyn4} & 59439 & 41607 & 17832 & 256 $\times$ 256 \\ 
        ~ & Red Lesion Endoscopy \cite{redlesionendoscopy} & 3895 & 2727 & 1169 & 320 $\times$ 320 \\ 
        ~ & FetReg \cite{fetreg} & 12334 & 8634 & 3700 & ~ \\ 
        ~ & TMEDOM \cite{tmedom} & 956 & 669 & 287 & ~ \\ 
        ~ & Positive Videos & 9496 & 6647 & 2849 & ~ \\ 
        US & COVID-19 Ultrasound \cite{covid19ultrasound} & 59 & 41 & 18 & ~ \\ 
        ~ & BUSI \cite{busi} & 780 & 546 & 234 & 500 $\times$ 500 \\ 
        ~ & DDTI \cite{ddti} & 134 & 94 & 40 & 560 $\times$ 360 \\ 
        ~ & MMOTU \cite{mmotu} & 1639 & 1147 & 492 & 330\~888 $\times$ 218\~657 \\ 
        ~ & HC18 \cite{hc18} & 1334 & 934 & 400 & ~ \\ 
        ~ & EchoNet-Dynamic \cite{echonet} & 10030 & 7021 & 3009 & 112 $\times$ 112 \\ 
        ~ & Positive Videos & 1874 & 1312 & 562 & ~ \\ 
        Dentistry & Panoramic radiography \cite{panoramicradiography} & 598 & 419 & 179 & 2041 $\times$ 1024 \\ 
        ~ & ODSI-DB \cite{odsidb} & 316 & 221 & 95 & ~ \\ 
        ~ & DENTEX 2023 \cite{dentex2023} & 2332 & 1632 & 700 & ~ \\ 
        ~ & Dental Calculus \cite{dentalcalculus} & 220 & 154 & 66 & ~ \\ 
        ~ & Vident-lab \cite{videntlab} & 15110 & 10577 & 4533 & 416 $\times$ 320 \\ 
        ~ & Dental condition \cite{oraldiseases} & 1296 & 907 & 389 & 612 $\times$ 408 \\ 
        ~ & Oral cancer \cite{oralcancer} & 144 & 101 & 43 & ~ \\ 
        ~ & Dental cavity \cite{dentalcavity} & 176 & 123 & 53 & ~ \\ 
        Surg & SARAS-ESAD \cite{sarasesad} & 27175 & 19023 & 8153 & 1920 $\times$ 1080 \\ 
        ~ & CholecSeg8k \cite{cholecseg8k} & 8080 & 5656 & 2424 & 854 $\times$ 480 \\ 
        ~ & DeSmoke-LAP \cite{desmokelap} & 6000 & 4200 & 1800 & ~ \\ 
        ~ & Surgical Hands \cite{surgicalhands} & 2838 & 1987 & 851 & ~ \\ 
        ~ & m2caiSeg \cite{m2caiseg} & 307 & 215 & 92 & 716 $\times$ 402 \\ 
        ~ & NeuroSurgicalTools \cite{neurosurgicaltools} & 2476 & 1733 & 743 & 612 $\times$ 460 \\ 
        ~ & ROBUST-MIS 2019 \cite{robustmis2019} & 10000 & 7000 & 3000 & 960 $\times$ 540 \\ 
        Optha & Cataracts \cite{cataracts} & 35127 & 24589 & 10538 & 1920 $\times$ 1080 \\ 
        ~ & Ocular Disease Recognition \cite{oculardiseaserecognition} & 3358 & 2351 & 1007 & 512 $\times$ 512 \\ 
        ~ & MeDAL Retina \cite{medalretina} & 2181 & 1527 & 654 & 768 $\times$ 768 \\ 
        ~ & RFMID \cite{rfmid} & 3200 & 2240 & 960 & 2144 $\times$ 1424 \\ 
        ~ & Glaucoma Detection \cite{GlaucomaDetection} & 650 & 455 & 195 & 3072 $\times$ 2048 \\ 
        ~ & DRIVE \cite{drive} & 40 & 28 & 12 & 584 $\times$ 565 \\ 
        Mammo & CBIS-DDSM \cite{cbisddsm} & 10239 & 7167 & 3072 & ~ \\ 
        ~ & CDD-CESM \cite{cddcesm} & 2006 & 1404 & 602 & 2355x1315 \\ 
        ~ & CMMD \cite{cmmd} & 5202 & 3641 & 1561 & ~ \\ 
        Genmed & LAION \cite{laion} & 10861 & 7603 & 3258 & ~ \\ 
        Non-medical & Celeb \cite{celeb} & 202599 & 60780 & 28364 & 178 $\times$ 218 \\ 
        ~ & Places \cite{places} & 10624928 & 637496 & 399497 & 200 $\times$ 200 \\ 
        ~ & AI2D \cite{ai2d} & 4903 & 3432 & 1471 & ~ \\ 
        ~ & DocFig \cite{docfig} & 33028 & 26422 & 6606 & ~ \\ 
        ~ & SciFig-Pilot \cite{scifig-pilot} & 263952 & 211162 & 52790 & ~ \\ 
        ~ & SlideImages \cite{slideimages} & 3452 & 2762 & 690 & ~ \\ 
        ~ & TextVQA \cite{textvqa} & 25119 & 20095 & 5024 & ~ \\ 
        ~ & SlideShare-1M \cite{slideshare1m} & 977605 & 782084 & 195521 & ~ \\ 
        ~ & Negative Videos & 23956 & 19165 & 4791 & ~ \\ 
        ~ & EgoHands \cite{egohands} & 4800 & 3840 & 960 & 720 $\times$ 1080 \\ 
        ~ & 11k Hands \cite{11khands} & 11076 & 8861 & 2215 & 1600 $\times$ 1200 \\ 
        ~ & IPN Hand \cite{ipnhand} & 95021 & 76017 & 19004 & 640 $\times$ 480 \\ \hline
    \end{tabular}
    \caption{Datasets used to train ResNet50 and ViT-Small medical image classifiers, used in Section \ref{sec:medical_video_filtering} and Section \ref{sup:pubmed_medical_filtering}.}
    \label{classifier-datasets}
\end{table*}

\begin{figure*}[ht!]
    \centering
    \includegraphics[width=\textwidth]{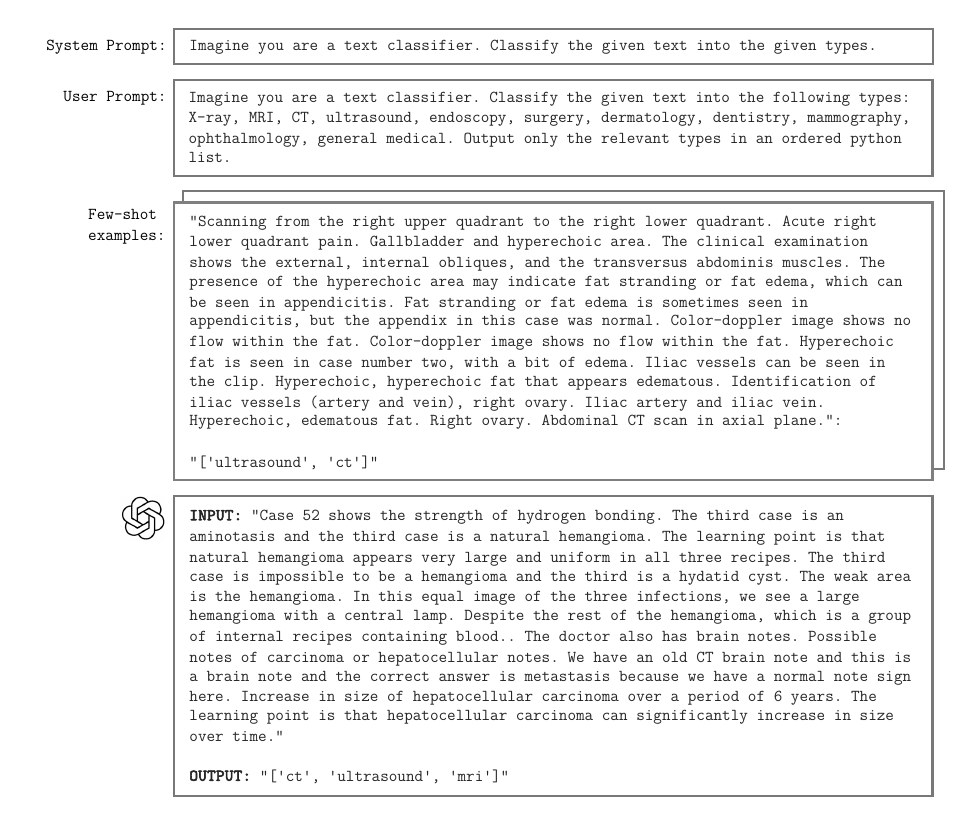} 
    \caption{The GPT-3.5 Turbo prompts used to determine whether a video contains discussion of multiple medical domains, with few-shot examples.}
    \label{fig:crossdomain-prompt}
\end{figure*}

\end{document}